\documentclass[journal]{IEEEtran}
\usepackage{graphicx}
\usepackage{float}
\usepackage{cite}
\usepackage{booktabs}
\usepackage{url}
\usepackage{subcaption}
\usepackage{amssymb}

\usepackage{tikz,xcolor,hyperref}

\definecolor{lime}{HTML}{A6CE39}
\DeclareRobustCommand{\orcidicon}{
\begin{tikzpicture}
\draw[lime, fill=lime] (0,0)
circle[radius=0.16]
node[white]{{\fontfamily{qag}\selectfont \tiny \.{I}D}}; 
\end{tikzpicture}
\hspace{-2mm}
}
\foreach \x in {A, ..., Z}{%
\expandafter\xdef\csname orcid\x\endcsname{\noexpand\href{https://orcid.org/\csname orcidauthor\x\endcsname}{\noexpand\orcidicon}}
}
\usepackage{tikz,xcolor,hyperref}
\definecolor{lime}{HTML}{A6CE39}
\DeclareRobustCommand{\orcidicon}{%
    \begin{tikzpicture}
    \draw[lime, fill=lime] (0,0) 
    circle [radius=0.16] 
    node[white] {{\fontfamily{qag}\selectfont \tiny ID}};    \draw[white, fill=white] (-0.0625,0.095) 
    circle [radius=0.007];    \end{tikzpicture}
    \hspace{-2mm}}
\foreach \x in {A, ..., Z}{%
    \expandafter\xdef\csname orcid\x\endcsname{\noexpand\href{https://orcid.org/\csname orcidauthor\x\endcsname}{\noexpand\orcidicon}}
    }

\ifCLASSINFOpdf
\else
\fi

\begin{document}
%

\title{Intelligent Amphibious Ground-Aerial Vehicles: State of the Art Technology for Future Transportation}

\author{Xinyu Zhang \hspace{-1.5mm}\orcidF{}, Jiangeng Huang \hspace{-1.5mm}\orcidD{}, Yuanhao Huang \hspace{-1.5mm}\orcidA{}
, Kangyao Huang \hspace{-1.5mm}\orcidC{}, Lei Yang\orcidG{}, Yan Han, Li Wang \hspace{-1.5mm}\orcidB{}, Huaping Liu \hspace{-1.5mm} \orcidE{}, \textit{Senior Member, IEEE}, Jianxi Luo\orcidL{} and Jun Li
\thanks{This work was supported by the National High Technology Research and Development Program of China under Grant No. 2018YFE0204300, and the National Natural Science Foundation of China under Grant No. U1964203, and sponsored by Meituan and Tsinghua University-Didi Joint Research Center for Future Mobility.}


\thanks{Xinyu Zhang, Jiangeng Huang, Lei Yang, Yan Han, Li Wang, and Jun Li are with the School of Vehicle and Mobility, Tsinghua University, Beijing, P.R.China (e-mail: xyzhang@tsinghua.edu.cn; e0575900@u.nus.edu; yanglei@mail.tsinghua.edu.cn; hanyan426926@163.com; wanglithu@mail.tsinghua.edu.cn; lj19580324@126.com)}%

\thanks{Yuanhao Huang is with the School of Vehicle and Mobility, Tsinghua University, Beijing, P.R.China, and also with the School of Aviation, Inner Mongolia University of Technology, Hohhot, Inner Mongolia, P.R.China (e-mail: huangyuanhao\_work@163.com)}%

\thanks{Kangyao Huang and Huaping Liu are with Department of Computer Science and Technology, Tsinghua University, Beijing, P.R.China (e-mail: kangyao.huang@outlook.com; hpliu@tsinghua.edu.cn)}%

\thanks{Jianxi Luo is the Founder and Director of Data-Driven Innovation Lab at Singapore University of Technology and Design (SUTD) (e-mail: luo@sutd.edu.sg)}%
}

\maketitle

\begin{abstract}
Amphibious ground-aerial vehicles fuse flying and driving modes to enable more flexible air-land mobility and have received growing attention recently. By analyzing the existing amphibious vehicles, we highlight the autonomous fly-driving functionality for the effective uses of amphibious vehicles in complex three-dimensional urban transportation systems. We review and summarize the key enabling technologies for intelligent flying-driving in existing amphibious vehicle designs, identify major technological barriers and propose potential solutions for future research and innovation. This paper aims to serve as a guide for research and development of intelligent amphibious vehicles for urban transportation toward the future.

\end{abstract}


\IEEEpeerreviewmaketitle

\section{Introduction}

\IEEEPARstart{M}{odern} transportation is a three-dimensional, comprehensive, and complementary network system consisting of vehicles of different types and modes. Discussion and attempts on urban three-dimensional transportation (3D transportation) and ``flying cars" are going on, such as the emerging Urban Air Mobility (UAM), roadable aircraft, and hybrid car-aircraft, \textit{et al}. In this work, we look at urban ground-air fusion transportation and amphibious ground-air vehicles. We first review the current state of amphibious vehicles. After that, we propose to introduce the intelligent system into the integrated fly-driving and summarize the enabling technologies that can be deployed in amphibious vehicles. These techniques include communication, perception, decision-making, as well as the intelligent control, with a special focus on safety improvement and energy saving. Moreover, barriers and potential solutions for intelligent amphibious vehicles and 3D transportation are listed.  Unlike other reviews, this work attempts to figure out what artificial intelligence technology can do to facilitate the vehicles in future 3D transportation. 

\par Various strategies have been proposed to achieve the urban 3D transportation. The most widely recognized view is that of UAM, a low-altitude transport that provides short-range flight service over the urban area. However, UAM is not flexible enough. It operates along predefined spatial corridors, so it cannot take full advantage of space. With further research, people pay attention to amphibious vehicles (roadway driving aircraft or flyable cars) that combine two motion modes and present the following advantages over vehicles with only one locomotion mode\cite{fan2019autonomous,qin2020hybrid,kalantari2014modeling,kalantari2013design,itasse2011equilibrium,elsamanty2012novel,sarica2019technology,sarica2021idea}:
\begin{itemize}
    \item Amphibious vehicles possess higher flexibility owing to multi-modal locomotion abilities.
    \item Ground-air transition mode can provide a better efficiency balance via a reasonable allocation between driving and flying.
    \item This is a real sense of fusion for 3D ground-air transportation.
\end{itemize}

\par Amphibious vehicles require land-air compatibility. For example, unique designs are required to enable both flying in the sky and driving on the roads. Similar techniques are frequently used in small-scale ground-aerial robots \cite{stoeter2002autonomous,lambrecht2005small,boria2005sensor,peterson2011wing,bachmann2009biologically,bachmann2009drive} to replace humans to perform operations in dangerous conditions. On the transportation level, manned flyable vehicles have appeared for many decades. However, 
the application is greatly restricted by technical difficulties. There are mainly three core techniques for amphibious vehicles: power technology, platform technology, and traffic management technology\cite{yangjun2020progress}. The power density and efficiency of power system directly determine the payload and endurance of vehicles. The power system should meet the requirements of vertical take-off and landing. Furthermore, platform technologies include high-performance configuration, lightweight design methods, and high adaptability approaches to ensure safety and land-air compatibility. In addition, how to build an efficient and secure low-altitude connected transportation system is an issue. This traffic system should be intelligent enough to support 3D transportation allocation and cloud scheduling.

\par In addition to the above techniques, flying and driving through cities need a complete intelligent mechanism to improve safety. It requires the resilience to failure to tackle uncertainties as well as unknowns of urban circumstance. Firstly, the system should be able to get a detailed awareness of its surroundings. Without sensing, vehicles can hardly provide protective responses amidst the uncertain and rapid-changing environment. With the progress of sensors, different kinds of radars play a vital role in helping vehicles perceive the environmental details. Moreover, planning and control are the next processes to ensure a comfortable and safe travel experience. Other information-level technologies involving self-locating, status sharing, communicating, and even autonomous piloting \textit{et al.,} are also indispensable parts of amphibious vehicles. In fact, a system including resilient detection, identification, localization, tracking, and airspace sharing is complicated enough. 

\par Considering the similarity between roadway autonomous driving and amphibious vehicle fly-driving, informatics techniques in roadway autonomous driving lay a solid foundation for safe fly-driving. However, it is impossible to directly transfer roadway intelligent technologies to amphibious vehicles without modification. Because the working conditions are different and flying itself is far more dangerous than driving on the ground, as well as the consequences of an accident are much more serious. Therefore, it is critical to expand ground intelligence and technologies into the spatial environment. For instance, to perceive the vehicle's state, to recognize environmental conditions, and to generate a reaction, as these capabilities fundamentally determine the autonomous system’s ability to fly safely\cite{Shish2021}. Recently, some enabling technologies have been applied to existing flyable vehicles that are discussed in detail in Section II\cite{Lombaerts2022}.

\par To achieve autonomous amphibious mobility and to improve safety, it is indispensable to make a comprehensive summary of intelligent techniques related to fly-driving. Prior review studies have been focused on different specific aspects of flying cars or 3D transportation like the prospect, crucial techniques, human perception, and ethics, \textit{et al}., but no one has yet reviewed the intelligent techniques used in UAM systems or amphibious vehicles involving perception, decision-making, as well as intelligent control. Herein, we collect the state-of-the-art literature related to amphibious vehicles to form a profile of this field. Since some prototypes and concepts have not been studied yet, filtered news and reports are selected as another source. Besides, research reports published by national governments are also within the scope of review. These sources are filtered strictly to avoid ambiguous and untruthful information. For instance, we only use physical test videos and prototypes as a criterion for media sources. To ensure academic authenticity and accuracy, concepts and hypothetical blueprint outlooks are distinguished.

\par Our work aims to review the state-of-the-art amphibious vehicles, and illustrates our insight of involving intelligent techniques into amphibious vehicles. In particular, we draw inspiration from the ground autonomous driving and autopilot flight technologies of UAVs to summarize the key characteristics of the autonomous fly-drive system. Then, we list the barriers and potential solutions of ground-air fusion 3D transportation.

\par The remainder of this paper is organized as follows. In Section \uppercase\expandafter{\romannumeral2}, existing amphibious vehicles are surveyed and summarized in the context of 3D urban transportation systems. Section \uppercase\expandafter{\romannumeral3} elaborates on the main points of artificial intelligence that could be used on amphibious vehicles in detail, from communication, near-ground perception, decision-making, and motion control. Section \uppercase\expandafter{\romannumeral4}  highlights the requirements and challenges for integrating amphibious vehicles into 3D urban transportation systems and provides potential solutions. The final section draws a conclusion.

\begin{table*}[]
\setlength{\abovecaptionskip}{0.3cm}

\caption{Statistics of Amphibious Vehicles}
\renewcommand\arraystretch{1.8
}
\centering
\setlength{\tabcolsep}{1mm}{
\scalebox{0.75}{
\begin{tabular}{ccccccccccc}
\hline
Year & Flying Car Name    & Organization                & \begin{tabular}[c]{@{}c@{}}Flight Mode\\ (HP/SP/Hybrid)\end{tabular} & Wing Type          & \begin{tabular}[c]{@{}c@{}}Can Wings\\ Be Folded\end{tabular} & Power Source       & \begin{tabular}[c]{@{}c@{}}Land and Air\\ Amphibious Function\end{tabular} & Country/Region & \begin{tabular}[c]{@{}c@{}}Maximum\\ Useful Load\end{tabular} & \begin{tabular}[c]{@{}c@{}}Maximum\\ range\end{tabular} \\ \hline
2009        & PAL-V Liberty      & PAL-V Company                & HP                                                                   & Rotor              & T                   & Hydrocarbon fuel & T                                                                          & Europe       & 246 kilograms                                                               & 500 kilometers                                          \\
2014        & Aeromobil 2.5          & Aeromobil             & HP                                                                   & Fixed wing       & T                   & Fuel             & T                                                                          & Slovakia      & 200 kilograms                                                        & 750 kilometers                                          \\
2015        & AirCar V5           & Klein-vision                 & HP                                                                   & Fixed wing       & F                   & Fuel             & T                                                                          & Slovakia     & 200 kilograms                                                    & 1000 kilometers                                         \\
2016        & Ehang 184          & Ehang Company                & SP                                                                   & Rotor              & F                   & Electricity      & F                                                                          & China        & 100 kilograms                                                    & 8.8 kilometers                                              \\
2017        & Joby S4            & Joby Aviation   & HP                                                                   & Rotor              & F                   & Electricity      & F                                                                          & America      & 381 kilograms                                                    & 320 kilometers                                          \\
2017        & Kitty Hawk Mk II   & Kitty Hawk Compa             & HP                                                                   & Fixed wing       & F                   & Electricity      & F                                                                          & America      & 115.2  kilograms                                                  & 160 kilometers                                          \\
2017        & Moller skycar M400 & Moller  & HP                                                                   & Fixed wing       & F                   & Fuel             & T                                                                          & America      & ——                                                                & 1200 kilometers                                         \\
2018        & Transition(TF-1)   & Terrafugia                & HP                                                                   & Fixed wing       & F                   & Hybrid     & T                                                                          & China        & 209 kilograms                                                    & 644 kilometers                                          \\
2018        & Volocopter         & Volocopter Company           & SP                                                                   & Rotor              & F                   & Electricity      & F                                                                          & Germany      & 160 kilograms                                                    & 35 kilometers                                           \\
2018        & Muyu eVTOL     & Muyu Aero Technology         & HP                                                                   & Fixed wing       & T                   & Electricity      & T                                                                          & China        & ——                                                   & 1500 kilometers                                         \\
2018        & CityAirbus         & Airbus Company               & SP                                                                   & Fixed   wing       & F                   & Electricity      & F                                                                          & Germany      & 250 kilograms                                                  & 80 kilometers                                           \\
2018        & Vahana Beta          & Airbus                       & Sp                                                                   & Fixed wing       & F                   & Electricity      & F                                                                          & Europe       & 200 kilograms                                                    & 100 kilometers                                              \\
2018        & BlackFly           & Opener                       & HP                                                                   & Fixed wing       & F                   & Electricity      & F                                                                          & America      & 90 kilograms                                                    & 30 kilometers                                          \\
2018        & Cora               & Wisk                         & SP                                                                   & Fixed   wing/Rotor & T                   & Electricity      & F                                                                          & New Zealand  & 181 kilograms                                                   & 100 kilometers                                         \\
2019          & Pegasus PAV        & Aurora Flight Sciences       & Hybrid                                                               & Fixed wing/Rotor & F                   & Electricity      & F                                                                          & America      & 225  Kilograms                                                    & 80 kilometers                                          \\
2019        & Ehang 216           & Ehang Company                & SP                                                                   & Rotor              & F                   & Electricity      & F                                                                          & China        & 200 kilograms                                                     & 35 kilometers                                         \\
2019        & Heaviside        & Kitty Hawk                   & HP/hybrid                                                            & Rotor              & T                   & Electricity      & F                                                                          & America      & ——                                                                & 160 kilometers                                          \\
2020        & TF-2A         & Terrafugia                & HP                                                                   & Fixed   wing/Rotor & F                   & Electricity      & F                                                                          & China        & 60 kilograms                                                    & 100 kilometers                                          \\
2021        & Maker Airtaxi      & Archer Company               & HP                                                                   & Fixed wing       & F                   & Electricity      & F                                                                          & America      & ——                                                                & 96 kilometers                                                   \\
2021        & Voyager X2           & XPeng                        & SP                                                                   & Rotor              & T                   & Electricity      & F                                                                          & China        & 200 kilograms                                                    & 35 kilometers                                      \\
2021          & V1500M             & AutoFlight  & SP                                                                   & Fixed wing/Rotor & F                   & Electricity      & F                                                                          & China        & Approximate 400 kilograms                             & 250 kilometers                                                      \\                                          \\ \hline
\end{tabular}}}
\label{tab1}
\end{table*}

\section{Amphibious Ground-Aerial Vehicles}
\par We start with a thorough review on amphibious ground-air vehicles. In this section, the background of amphibious vehicles is described firstly. Then, we select four representative vehicles that have been put into practice and illustrate them in details. It needs to be emphasized that the supporting materials come from multiple sources including media, reports, and academic papers. Besides, features of amphibious vehicles are summarized and differences from UAM vehicles are also listed.

\subsection{Background}
\subsubsection{Social Context}

\par Ground-aerial fusion transportation gives the public enduring appeal since the birth of aircraft. According to Deloitte, ``Future Flying Cars for Mobilit" is expected to create a 13.8-billion-dollar market in 2040 in the US\cite{Garrow2021}. In 2018, Morgan Stanley predicted that the market size of autonomous flying cars may reach 1.5 trillion US dollars in 2040. National Aeronautics and Space Administration (NASA) has launched Advanced Air Mobility mission in 2021 to transport people and cargoes using revolutionary aircraft where was totally not served or underserved by aviation previously\cite{nasaAdvancedMobility}. Besides, the European Urban Air Mobility initiative, which aims to connect companies, regions, and cities together, and has gained support from the European Commission. Recently, China also proposed the blueprint of ground-aerial transport. China's Ministry of Transport and the Ministry of Science and Technology jointly issued the ``Medium- and long-term development Plan for Scientific and technological innovation in transportation (2021-2035)", which clearly pointed out the potential of fly-driving technologies, including amphibious mode switching, and the integration of aircraft and automobiles\cite{ChinasMinistryofScienceandTechnology2022}.

\subsubsection{History}
\par
\begin{figure}[h]
\centering
\includegraphics[width=3.5in]{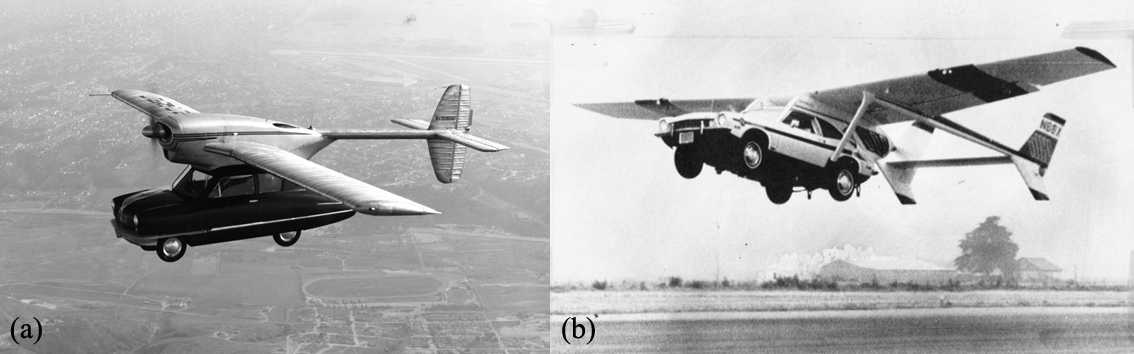}
\caption{ \centering
Earlier flying car. (a) ConvAirCar Model 118 (1947). (b) AVE Mizar (1971).
(\url{https://www.vintag.es/2018/08/flying-cars.html})\cite{VINTAGEEVERYDAY2018}}
\label{earlier flying car}
\end{figure}

 Since airplanes were created, inventors started their crazy experiments about flying cars. The initial concept and prototype made an accurate interpretation of the term ``flying car". Most of the earlier amphibious ground-aerial vehicles are the simple combination of airplane and automobile. Original flying cars had fixed wings that were mounted to automobiles\cite{Pan2021}. Here we present two typical prototypes of earlier flying car in Fig.\ref{earlier flying car}. The situation continues until 1960s, rotary wings were introduced into amphibious vehicles. Amphibious vehicles obtain ability to take-off and land vertically with rotary wings. Piasecki VZ-8 AirGeep is a well-known example, with two huge vertical propellers at the front and back of the vehicle, a pilot seat in the middle, and three or four wheels for roadway driving\cite{Pan2021}. However, this attempt was eventually abandoned. Indeed, due to the lack of public acceptability and the immaturity of the technologies, the majority of past attempts to build flying automobiles failed.
    
\par Recently, the electricity-driven vertical take-off and landing technology for heavy-loaded UAVs and aircrafts is developing rapidly; meanwhile, intelligent technologies have been used successfully. Due to the maturity in the high energy density battery and breakthrough of electric drive technology, amphibious vehicles usher explosive growth since 2010s. Nowadays, aviation giants including Airbus, Boeing, Bell, traditional car companies such as Audi, Toyota and Geely, and uprising technology companies such as Uber from the US, Ehang from China, and Volocopter in Germany \textit{et al.} have been prototyping flying cars. With these progress, the confidence of investors, governments and industrial companies is growing\cite{karnouskos2018self}.

\subsection{Representative Amphibious Vehicles}

\par We survey academic papers and online information after 2009 to summarize the state of existing amphibious vehicles. Some companies may have more than one version, here we only count the first manned test version. The results are shown in Table.\ref{tab1}. Presently, no technical and functional standard exists for flying cars worldwide, whereas various taxonomies have been proposed. For example, according to the way thrust is generated, vehicles can be divided into fixed-wings and rotary-wings. According to power sources, vehicles can be divided into electric-driven and fuel-driven types. In addition, by the category of actuators, they can also be divided into fixed-rotor and tilt-rotor. Hence cases are arranged in chronological order.

\par So far amphibious vehicles develop into a complex system. Its design requires the integration of knowledge from different disciplines, such as material science, dynamics, mechanics, and human-machine interaction design. Interdisciplinary research and development are needed to solve complicated, highly coupled problems among modules under numerous constraints. According to the summary above, some prototypes have matured to the point where they can be purchased, such as the AirCar, Aeromobil, Transition TF-1, and PAL-V Liberty.

\begin{figure*}[t]
\centering
    \begin{subfigure}{.3\textwidth}
    \includegraphics[height=1.2in]{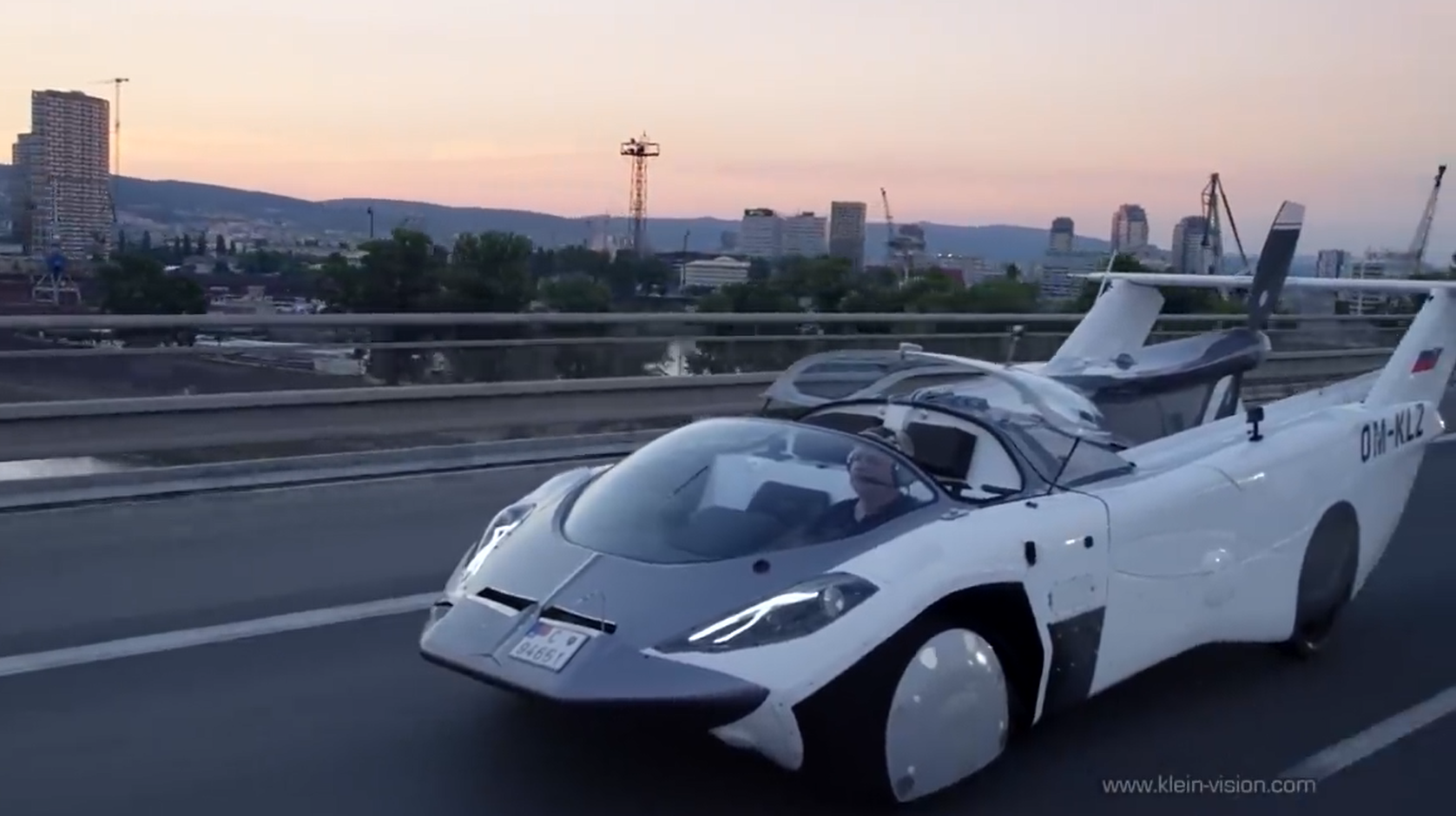}
    \caption{Roadway driving.}
    \end{subfigure}
    \begin{subfigure}{.3\textwidth}
    \includegraphics[height=1.2in]{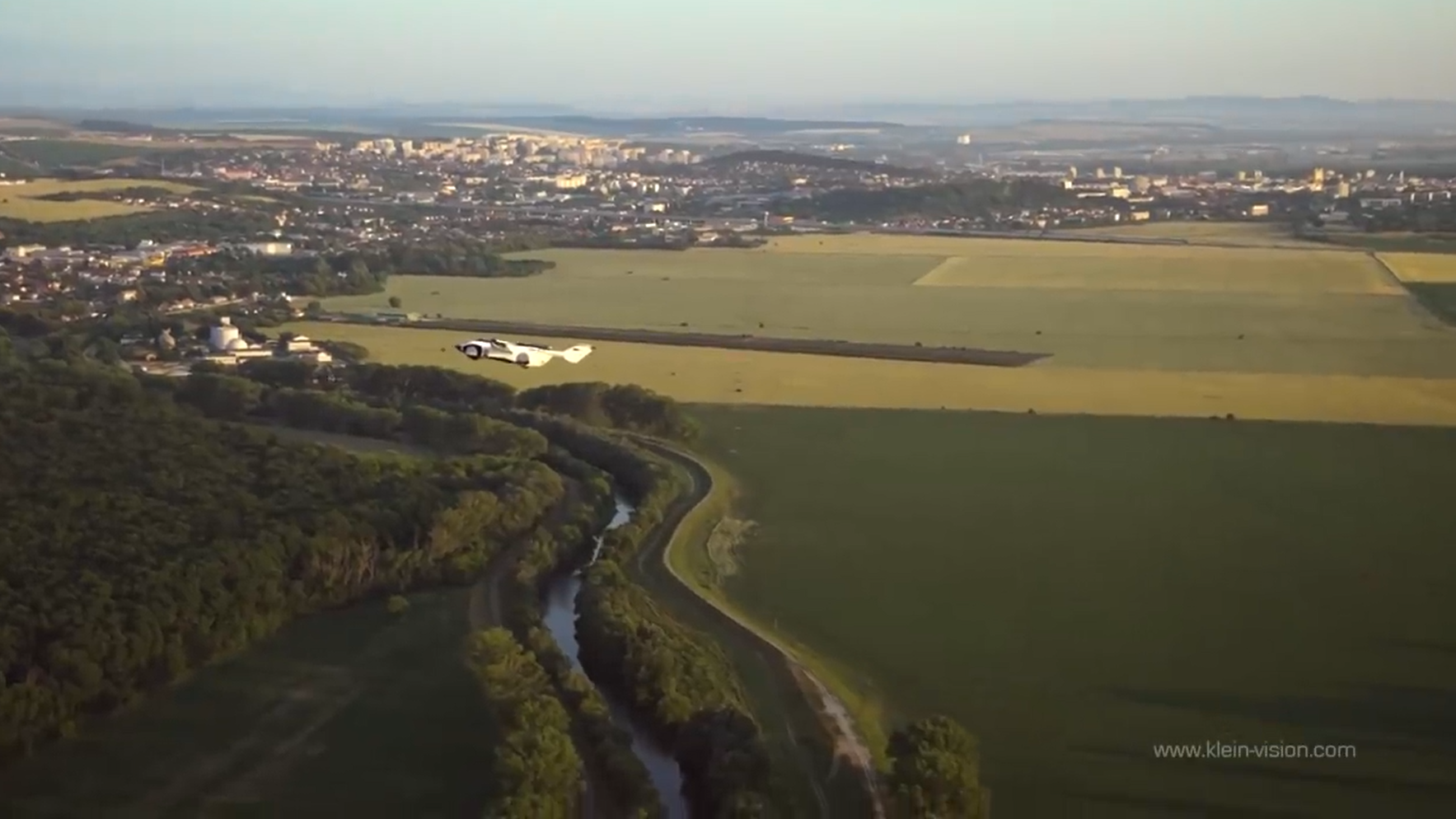}
    \caption{Flying over cities.}
    \end{subfigure}
    \begin{subfigure}{.3\textwidth}
    \includegraphics[height=1.2in]{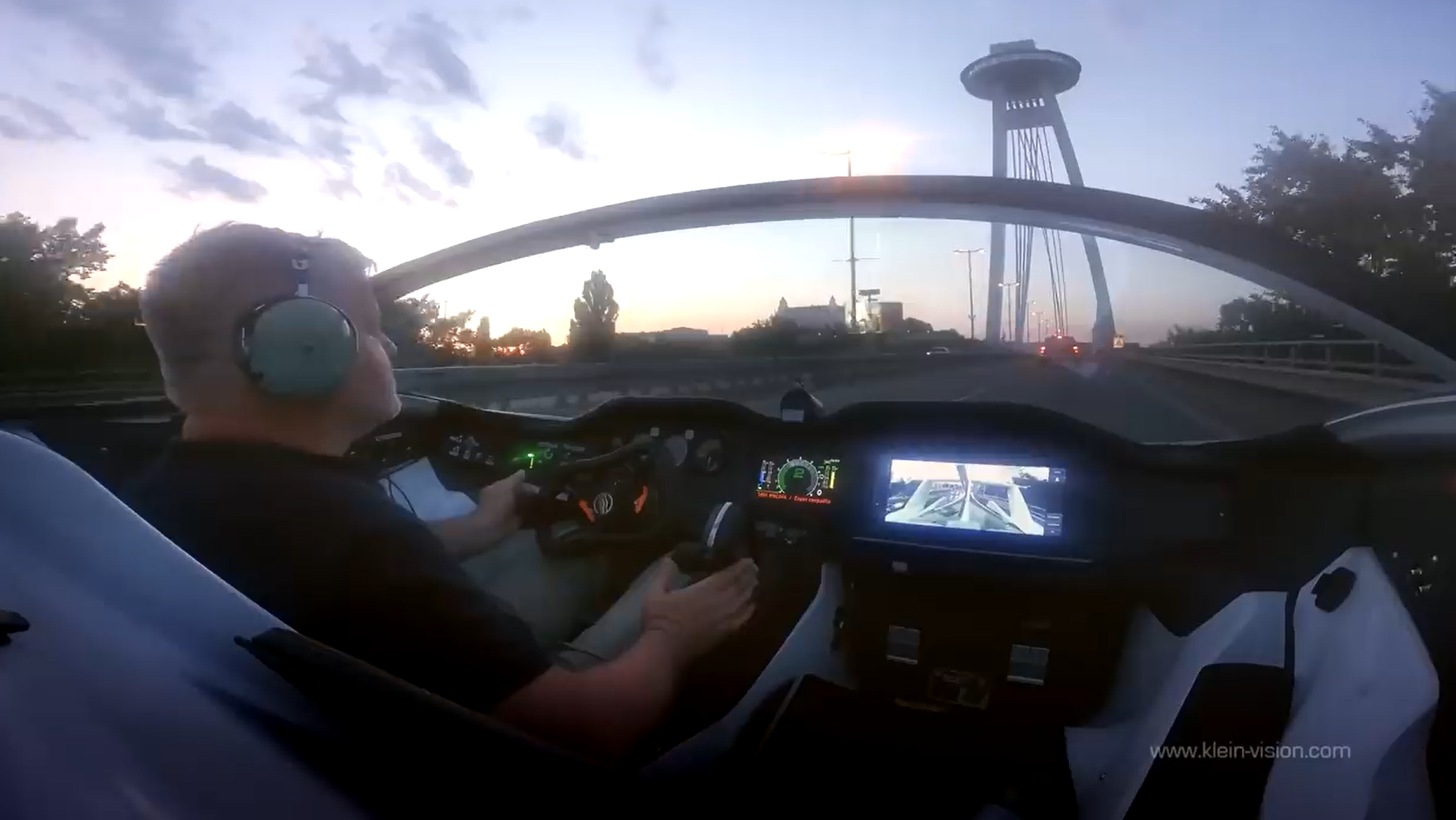}
    \caption{Integrated console.}
    \label{klein vision console}
    \end{subfigure}
    \caption{The hybrid car-aircraft AirCar by KleinVision. \url{https://www.klein-vision.com/}}
    \label{klein vision}
\end{figure*}

\begin{figure*}[t]
\centering
    \begin{subfigure}{.3\textwidth}
    \includegraphics[height=1.2in]{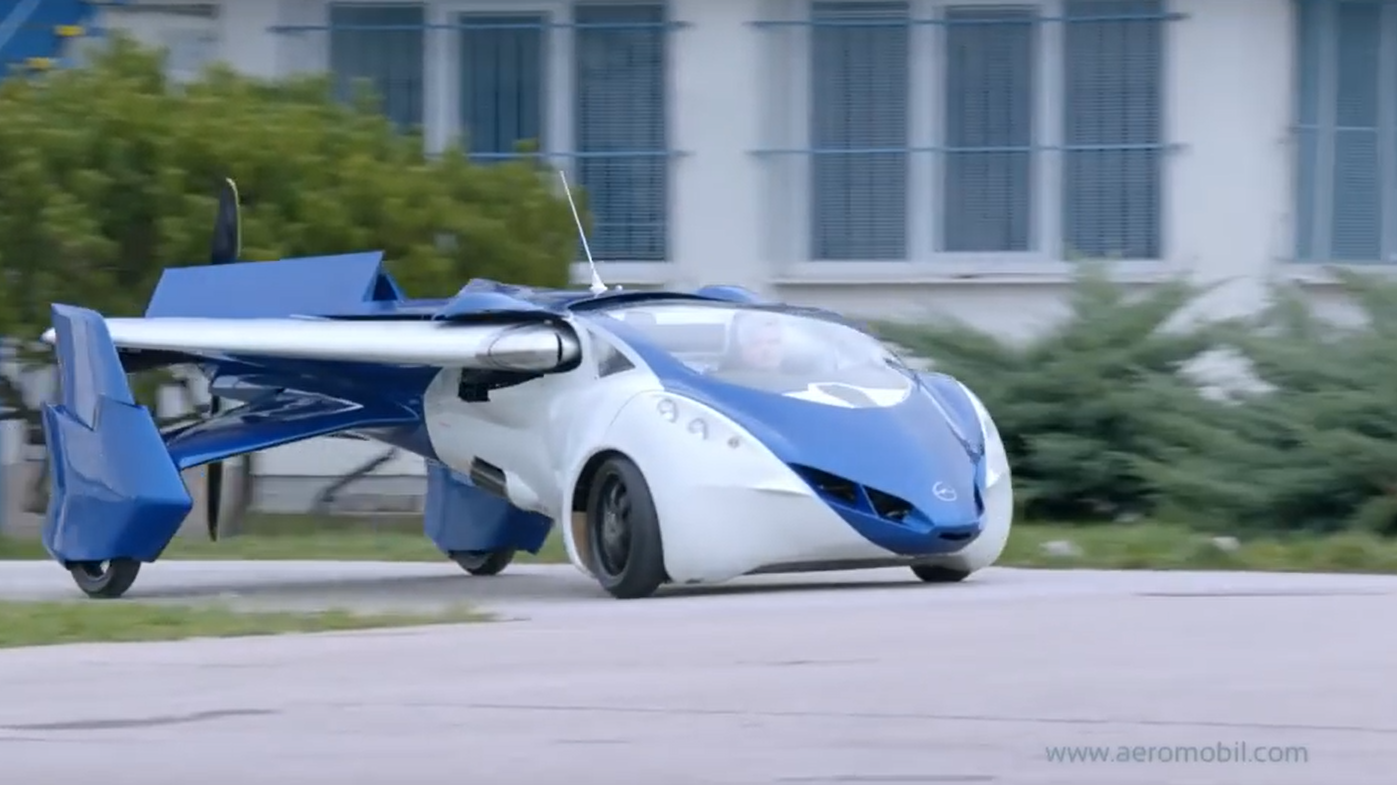}
    \caption{Roadway driving.}
    \end{subfigure}
    \begin{subfigure}{.3\textwidth}
    \includegraphics[height=1.2in]{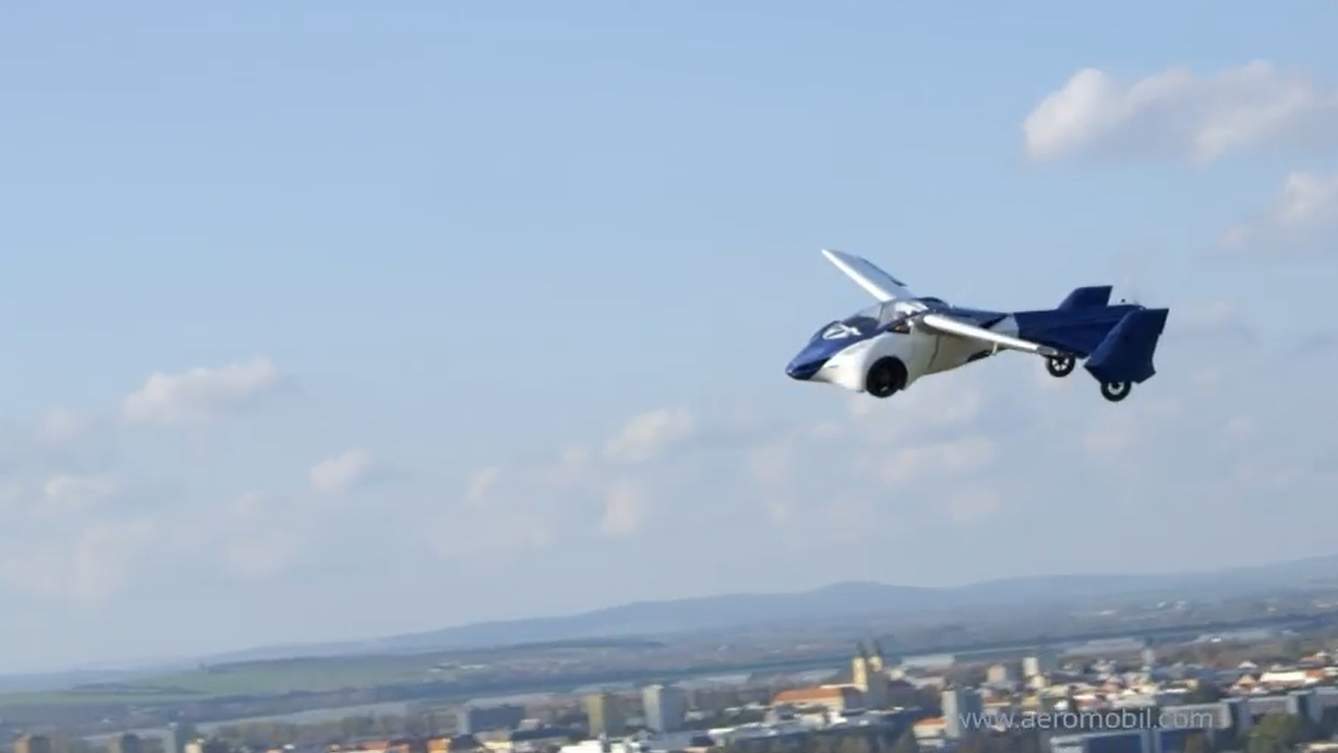}
    \caption{Flying over cities.}
    \end{subfigure}
    \begin{subfigure}{.3\textwidth}
    \includegraphics[height=1.2in]{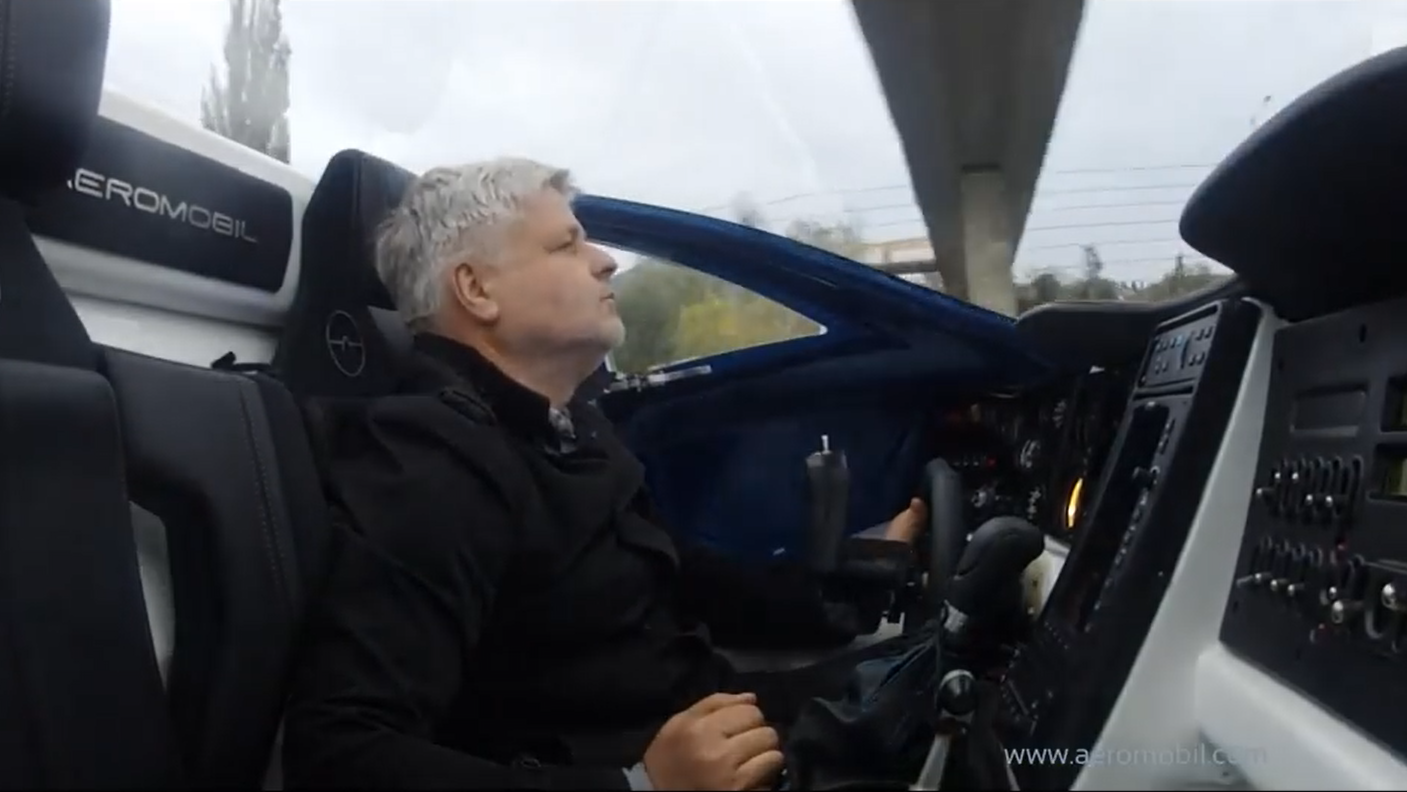}
    \caption{Integrated console.}
    \label{aeromobil console}
    \end{subfigure}
    \caption{The amphibious flying car Aeromobil.
    \url{https://www.aeromobil.com/}}
    \label{aeromobil}
\end{figure*}

\begin{figure*}[t]
\centering
    \begin{subfigure}{.3\textwidth}
    \includegraphics[height=1.2in]{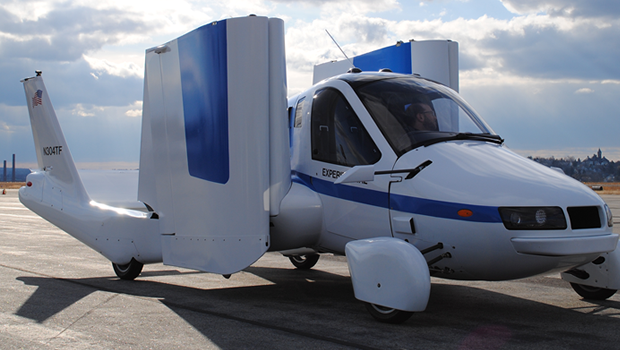}
    \caption{Roadway driving.}
    \end{subfigure}
    \begin{subfigure}{.3\textwidth}
    \includegraphics[height=1.2in]{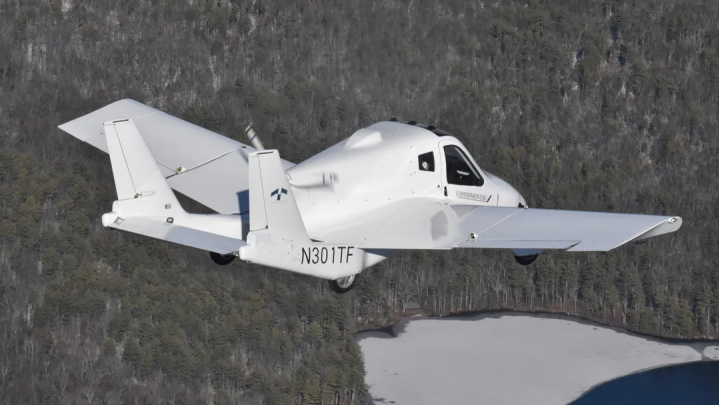}
    \caption{Flying over cities.}
    \end{subfigure}
    \begin{subfigure}{.3\textwidth}
    \includegraphics[height=1.2in]{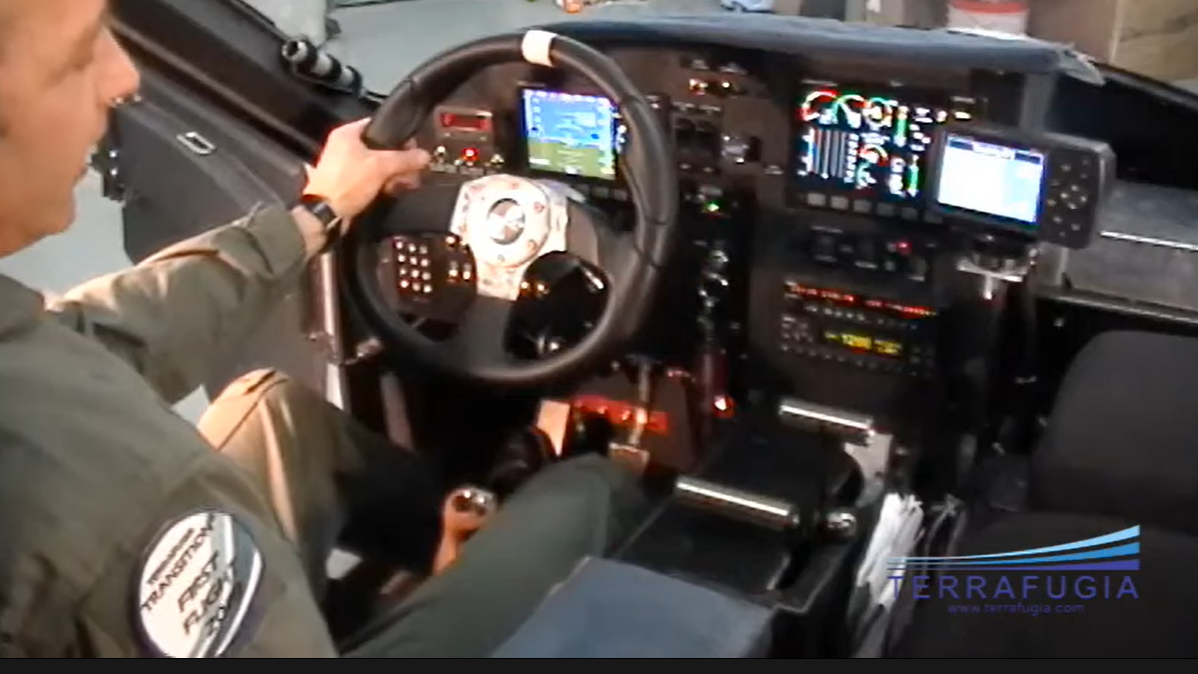}
    \caption{Inside the vehicle.}
    \label{TF-1 console}
    \end{subfigure}
    \caption{Roadable vehicle Transition (TF-1) developed by Terrafugia.
    \url{https://terrafugia.com/transition/}}
    \label{TF-1}
\end{figure*}

\begin{figure*}[t]
\centering
    \begin{subfigure}{.3\textwidth}
    \includegraphics[height=1.2in]{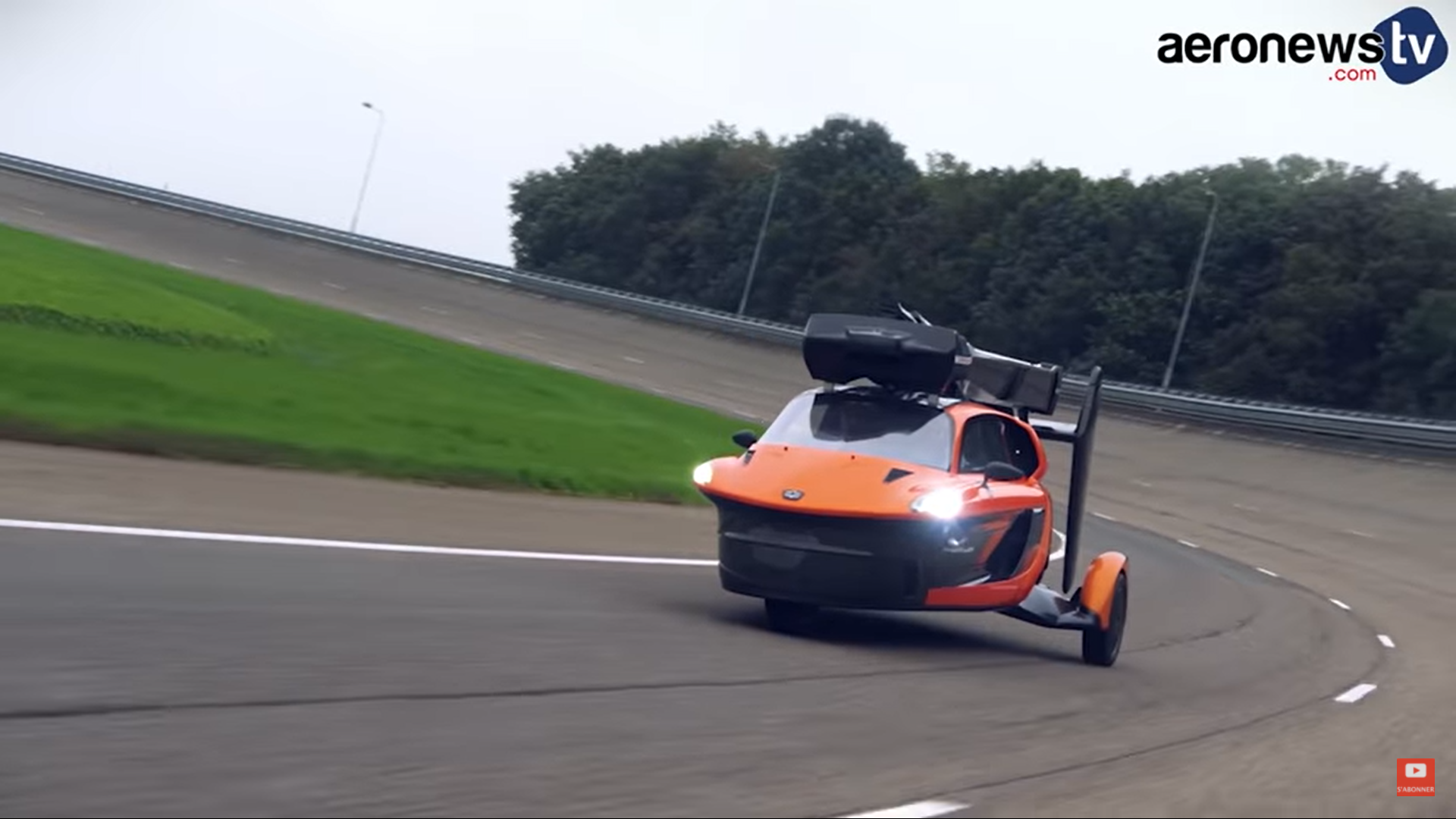}
    \caption{Roadway driving experiment.}
    \end{subfigure}
    \begin{subfigure}{.3\textwidth}
    \includegraphics[height=1.2in]{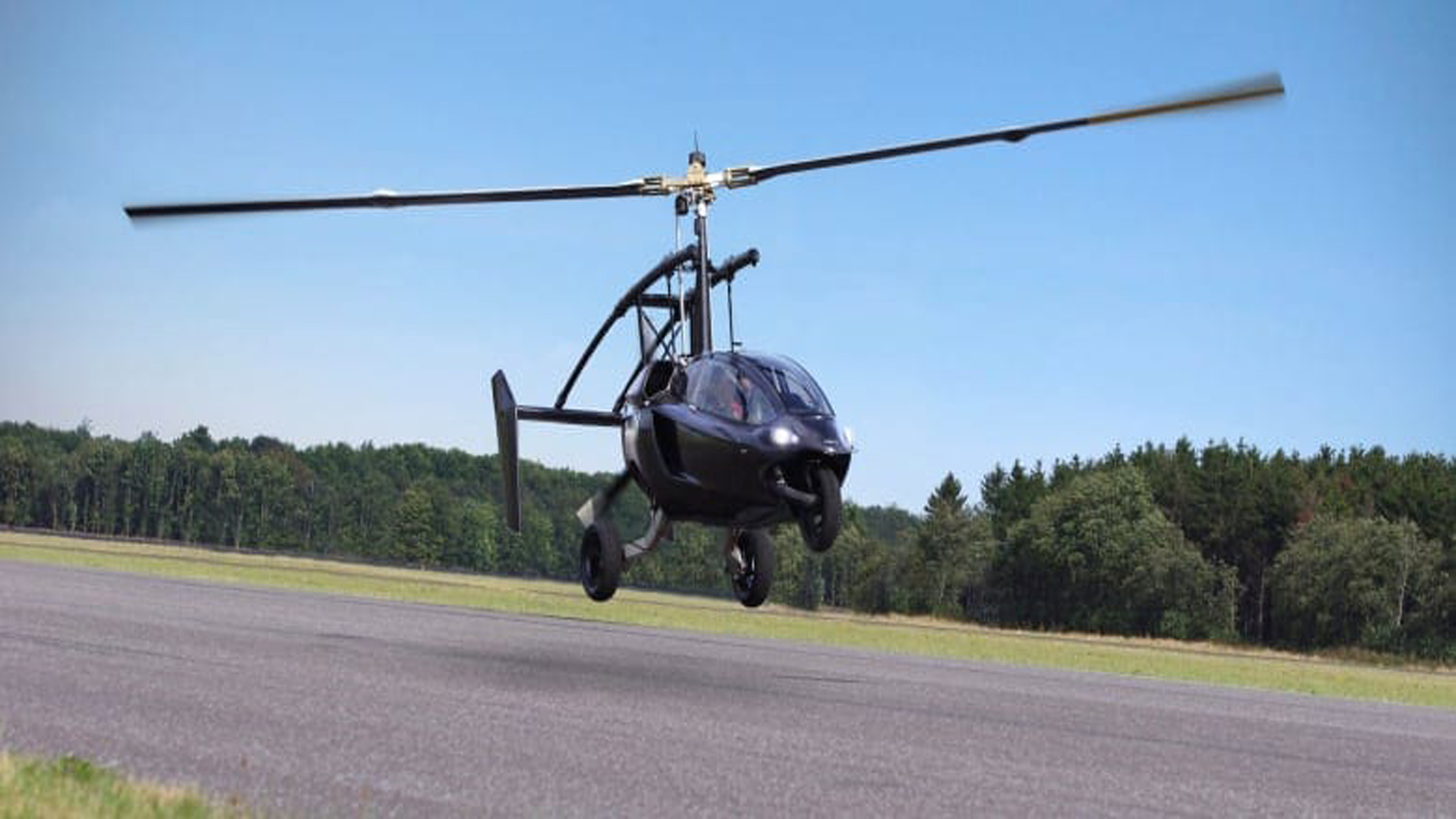}
    \caption{Flying experiment.}
    \end{subfigure}
    \begin{subfigure}{.3\textwidth}
    \includegraphics[height=1.2in]{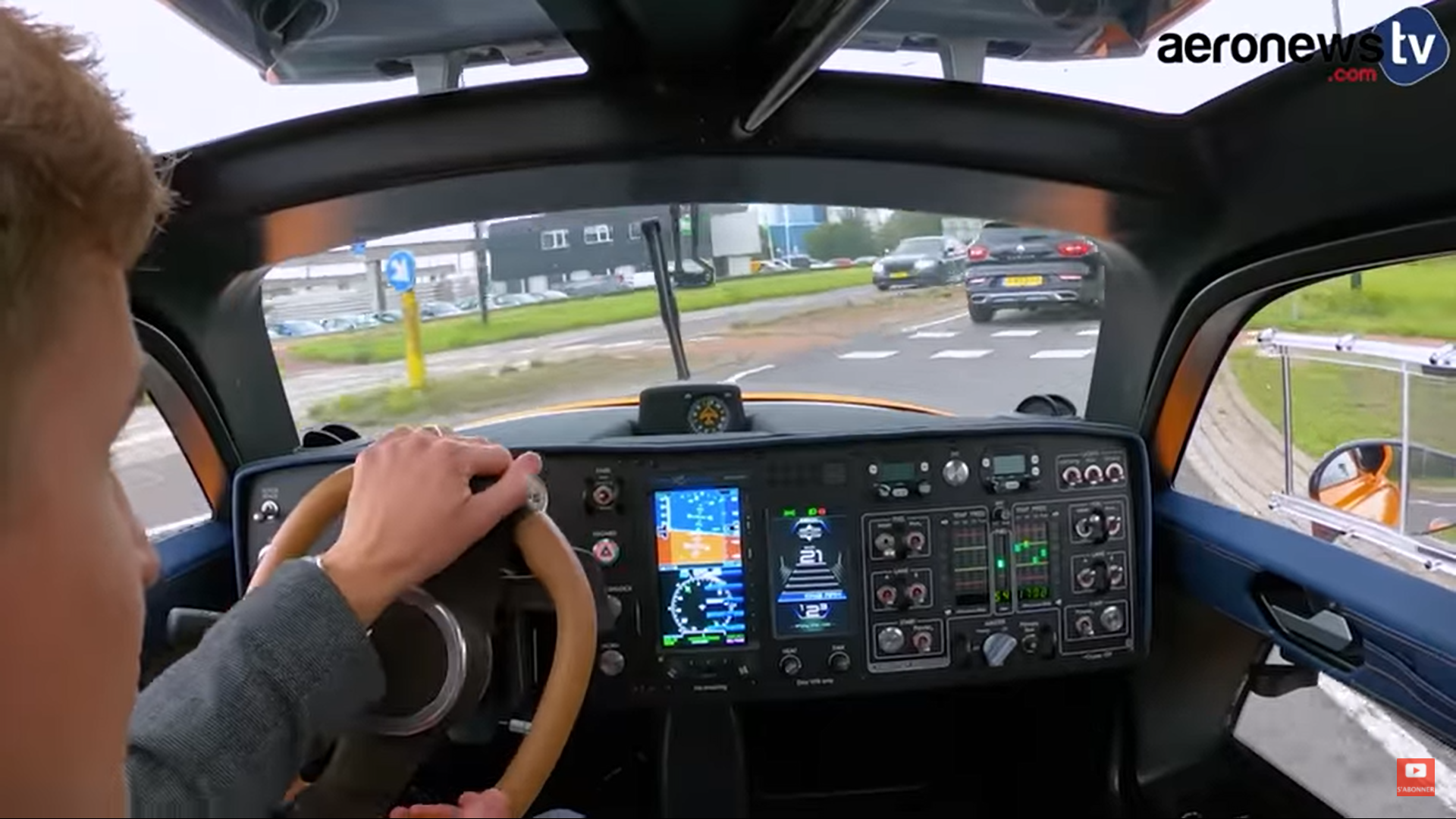}
    \caption{Inside the vehicle.}
    \label{PAL-V console}
    \end{subfigure}
    \caption{Amphibious vehicle by PAL-V Liberty.
    \url{https://www.pal-v.com/}
    \newline 
    \centering
    \url{https://www.aviationtoday.com/2021/02/23/flying-car-receives-certification-basis-easa/}}
    \label{PAL-V}
\end{figure*}

\par Prototypes with fixed-wings are more prevalent. For instance, AirCar is an amphibious vehicle designed and manufactured by KleinVision, shown in Fig.\ref{klein vision}. This prototype has completed a 35-minute flight between international airports in Nitra and Bratislava, Slovakia, last year. This vehicle can carry two people, with combined weight limit of 200kg\cite{Kleinman2021}. In addition, as shown in Fig.\ref{aeromobil}, Aeromobil is a multi-modal flyable vehicle that is similar to AirCar. This version 2.5 had completed its flight trial for roughly 40 hours as early as 2014\cite{PeterSigal2014}. Furthermore, Transition (TF-1) is a light roadable vehicle developed by Terrafugia, owned by Geely Holding Group, China, as shown in Fig.\ref{TF-1}. To drive on the road, all three of these feature collapsible fixed wings. Aeromobil and KleinVision's wings can be folded backwards, while TF-1's wing is segmented and can be folded to the sides.

\par PAL-V Liberty is the representative of rotary-wings amphibious vehicle, shown in Fig.\ref{PAL-V}, which can carry more than 240 kilograms load. Although PAL-V's rotary-wing reaches 10.75m, its rotor mast can fold automatically when driving. The most noteworthy is the PAL-V can run on regular fuel. Other rotary-wing amphibious vehicles are not mature enough for commercial use. For example, Pop.Up by Airbus proposes a trailblazing modular ground and air passenger concept vehicle system, dividing a vehicle into three modules \cite{airbusItaldesignAirbus}. A passenger capsule is designed to be coupled with two different and independent electrically-propelled modules (the ground module and the air module), and both of them can be separated from the passenger capsule. Similar idea can also be found on ``FulMars" from Cool Black Technology.

\subsection{Typical Features of Amphibious Vehicles}
Here we conclude typical features of amphibious vehicles into two points: transition between multi-modal locomotion, and integrated driving-flying system.

\subsubsection{Transition between Flying and Driving Modes}
\par The transition techniques of amphibious vehicles are diverse\cite{elsamanty2012novel,Salman2020}. By runways is the most frequent method for transition since majority of amphibious vehicles adopts fixed-wings to generate lifting force. These type of vehicles have classic chassis for land driving and complete avionics systems for flight. To overcome gravity, sufficient airspeed is required to generate lifting force\cite{fadhil2018gis}. Thus, they show more similarities to conventional aircrafts and are more sensitive to landing conditions\cite{pardede2020take}. Currently, fixed-wings based vehicles use airports for ground/air transition. Some researchers have explored novel configuration and controllers to shorten the required take-off or landing distance while still hard to be used in cities.

\par Consequently, vertical take-off and landing (VTOL) technology make sense for urban ground/aerial transportation. VTOL is less influenced by the surroundings, has greater flexibility\cite{Postorino2020}, and more fitness when operating in complex and unknown environments\cite{metinelectrically}. This technique is originally derived from unmanned aerial vehicle. Current studies on VTOL vehicles are generally based on the fundamentals of multi-rotors and tilt-rotors. Presently, numerous vehicles adopt the VTOL pattern\cite{rajashekara2016flying}, such as the ``FulMars" from Cool Black Technology Co., Ltd., the ``Pop.Up Next" developed in cooperation between Audi and Airbus, the ``Pegasus" flying vehicle from Pegasus International Inc., and the ``Black Knight" land-air amphibious cross-country vehicles developed by the support of the U.S. Army. In general, vertical lifting requires a much higher power density than cruising.

\subsubsection{Integrated Avionics and Control System}
\par Furthermore, the integration of the ground driving console with the avionics system is critical. This characteristic only exists in amphibious ground-aerial vehicle. Generally, avionics provides protection for flight. The avionics installed in an aircraft or spacecraft can include engine controls, flight control systems, navigation, communications, flight recorders, lighting systems, threat detection, fuel systems, electro-optic (EO/IR) systems, weather radar, performance monitors, and systems that carry out hundreds of other mission and flight management tasks\cite{BAESystems}. Since amphibious vehicles owns multi-modal locomotion capabilities, avionics system should be integrated with the driving controller. As shown in Fig.\ref{klein vision console} \ref{aeromobil console} \ref{TF-1 console} \ref{PAL-V console}, avionics systems have been integrated into the car console, and the wheel is commonly employed in both flying and driving modes. 

\subsection{Differences from UAM Vehicles}
\par As aforementioned, amphibious vehicles have several advantages compared with the UAM vehicles, although UAM system is the mainstream in 3D fusion transportation research studies. Here, we illustrate the differences between these two categories.

\par UAM belongs to Advanced Air Mobility (AAM). With the development of VTOL, corridor services are envisioned for the 2020s \cite{Cohen2021}. Though their commercialization is still in the development stage and face several barriers in society problem like weather restrictions, and legal resistance, etc\cite{goyal2018urban}, it is still undeniable that they have great potential. One of the relevant studies believes that AAM would generate the annual market with a total value of 2.5 billion USD and carry almost 80 thousand passengers only in the U.S.\cite{goyal2021advanced}. With the mature of facilities, the UAM and other vertical mobility (VM) could serve as an brand new means of vehicles in the short-term intercity transportation and replace the train, car and plane \cite{grandl2018future}. Together with electrification and vertical take-off and landing ability\cite{wang2014autobody}, UAM becomes one of possible modes that offers high-efficient aerial short-term fixed-route service within or between cities\cite{Straubinger2020}. Despite of the shortcoming in flexibility, UAM has been tested as demonstration. The flying air taxi eHang 184 is an example of UAM that had been approved for testing in Nevada and actually experimented in Dubai\cite{BBC}.

\par Although both amphibious vehicles and UAM vehicles have the flying ability, they adopt unidentical working patterns. The core distinction is amphibious vehicles are both roadable and flyable. Compared with UAM vehicles, amphibious vehicles with dual mobility modes achieve a breakthrough to meet the increasingly complex mobility demands for high-density urbanization development. More disparities between UAM and amphibious vehicles could be found bellow:

\subsubsection{Essence}
\par Amphibious vehicles are flyable roadway vehicles that have most of the functions of a conventional vehicle. UAM vehicles, on the other hand, are essentially aeroplanes in cities.

\subsubsection{Flexibility}
\par Theoretically, amphibious vehicles are more flexible. As aforementioned, the UAM strategy is a fixed-route fusion method for urban air transport, whereas amphibious vehicles can transfer between multiple modes under the discretion of the driver. In the UAM system, public take-off and landing demands physical platforms for the intersection between land and air \cite{2020An,luo2021simulation,silva2018vtol}, and passengers should be transferred at these fixed stations. 

\subsubsection{Functional Orientation}
\par Amphibious vehicles are typically used for personal transportation, whereas UAM is for public use to transport people and freight. So far, most amphibious vehicles can carry less than two passengers. UAM vehicles can carry more in general since they do not require a chassis with a huge load capacity. For example, electric VTOL aircraft like the S4 by Joby Aviation and the V1500M manufactured by AutoFlight can lift four passengers.

\subsubsection{Traffic Management Strategy}
\par Amphibious vehicles and UAM vehicles require different strategies for effective fusion in urban transportation system. Compared with amphibious vehicles, UAM vehicles have no need to blend into ground traffic. The traffic management for UAM may be totally tiered by height without interfering with ground traffic because it serves along fixed airlines over cities. 

\section{Intelligent Technologies to Facilitate Amphibious Ground/Aerial Vehicles in 3D Transportation}

\par In traditional ground transportation, intelligent assisted systems like perception and emergency decision modules can make driving easier and enhance driving safety. For example, functions like lane-keeping, adaptive cruise control (ACC), and emergency obstacle avoidance are applied in most cars. In addition, the autopilot provides a high-level self-driving function to relieve driving fatigue. The aforementioned examples prove that vehicles can benefit from the intelligence system, especially from the perspectives of driving safety and autonomous ability. 

\par Meanwhile, applying intelligent technologies to flyable vehicles is showing great appeal for developers. For example, \cite{Vempati2021} looks at the human-automation role in 3D transportation, and \cite{Kelley2022} gives a scenario of autonomous flyable vehicles in urban areas. However, fly-driving in 3D fusion transportation, which includes multi-modal locomotion, is more complicated than ground driving. This complexity makes greater demands on intelligent techniques than those in traditional ground autonomous driving.

\par As aforementioned features of amphibious vehicles, we can extract the advantages of introducing intelligent system into 3D transportation:

\begin{itemize}
    \item Firstly, intelligent technologies make amphibious vehicles more available for individuals. The implementation of an intelligent control system can significantly reduce the cost of pilot training. In addition, it can maintenance and simplify tedious procedures, which lowers the barriers for starters.
    
    \item Secondly, intelligent autonomous technologies can make fly-driving safer. A computer-aided driving system can detect and alert obstacles, and usually has more accurate results than a human-operating system, such as a mandatory collision avoidance system for obstacles. 
    
    \item Thirdly, the intelligent system benefits traffic management. It can reduce administrative costs and errors, and reduce human intervention. To achieve more effective regulation, an intelligent connected system and other appropriate types of equipment should be adopted, ensuring adequate guidance and supervision, and even a remote mandatory takeover.

\end{itemize}

\par Therefore, intelligent systems should be introduced to increase the safety of both flying and driving modes. Amphibious vehicles need a powerful manual and intelligent auto-pilot system for both manned and unmanned modes. The intelligent system should have accurate environmental perception, robust decision-making and planning strategies, and highly redundant controllers. In addition, other issues like safety and emergency handling must be considered.

\par
\begin{figure*}[htbp]
  \centering
  \includegraphics[width=0.8\linewidth]{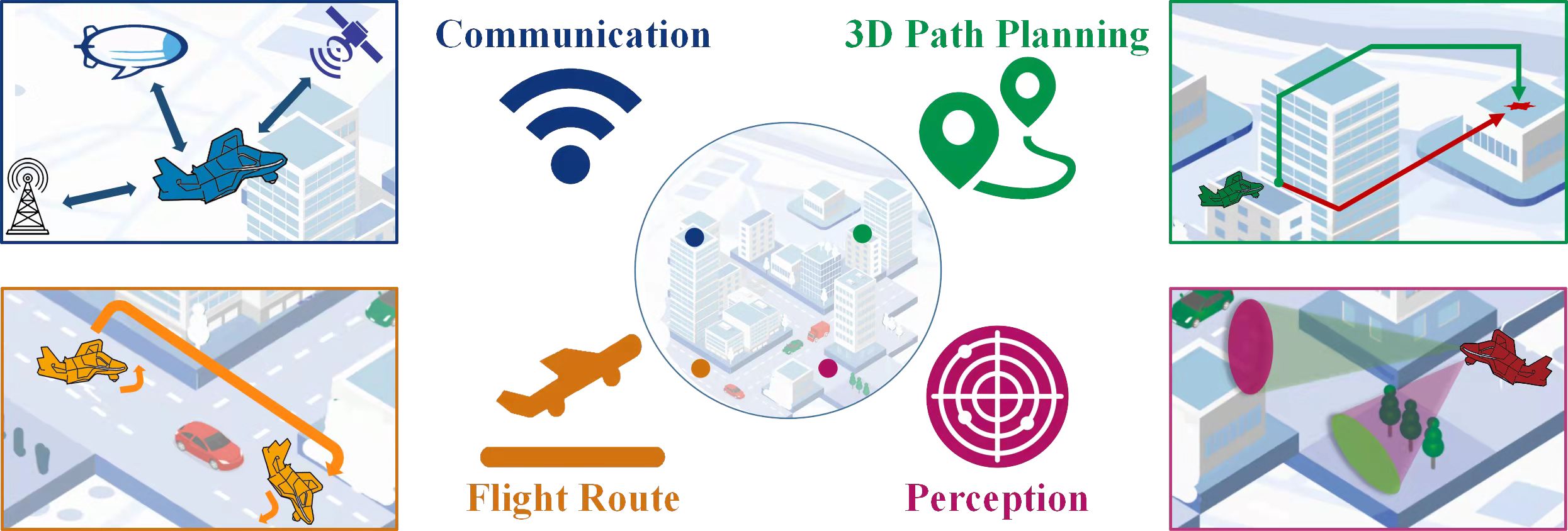}
  \caption{Main intelligent technologies to facilitate the 3D transportation.}
  \label{overview}
\end{figure*}

 Herein, we consider the intelligent technologies for amphibious or any other types of flyable vehicles from several sub-categories, similar to the division of on-road self-driving technology: intelligent connected systems, perception, decision-making and path planning, and intelligent control. Main intelligent technologies to facilitate 3D transportation are shown in Fig.\ref{overview}:

\begin{itemize}

\item The connected system, which serves as the infrastructure facility for intelligent 3D fusion transport, should be robust, secure, and stable for vehicles in 3D transportation. The connected system is far more than a communication module. It should have not only information sharing but also cooperation functions so as to build a solid foundation for intelligent traffic management.

\item The perception system serves as the eyes for intelligent amphibious vehicles. The perception module should capture the surrounding information and process it promptly, then send the key information to the decision-making module. More specifically, for a better driving experience, the vehicle should pay attention to buildings and other impediments throughout the near-ground flight and issue assistance orders as needed. Additionally, perception can offer more precise information for takeoff and landing, particularly for the space where eyesight is limited.

\item In both fly-driving and intelligent 3D fusion transportation, the decision-making module acts as the brain of the vehicle. The optimal route can be automatically chosen by vehicles to avoid the obstructions. In 3D transportation, obstacles include nearby structures, other low-flying vehicles, and hung items at low altitudes. Thus, planning is challenging due to the rise in spatial complexity.

\item The intelligent control systems promote the mobility of amphibious vehicle. The system is anticipated to perform the duties of the ground-based driving control as well as the needs of the aerial autopilot system, including attitude stabilisation and locomotion control. Furthermore, when flying close to the ground, the ground effect by flow field disruption must be considered.

\end{itemize}

\par The rest of this section is organized by these four categories of core techniques.

\subsection{Intelligent Connected System}

\par The 3D transportation system is a classic multi-agent model. Numerous vehicles flying in the sky with chaotic trajectories could be a catastrophe. Management regulations should be performed by a real-time command and control system, otherwise the individuals should be equipped with sensing and transmitting devices to make contact with their surroundings. Thus, regardless of the form of communication, either decentralized or centralized, the communication system is crucial. The connected system should be concerned with communication loss and communication delay and should always be robust, safe, and accessible for surrounding vehicles\cite{erturk2020requirements}.

\par We divide the connected system into two parts: vehicle-to-vehicle (V2V) communication and vehicle-to-infrastructure (V2I) communication. In contrast to centralised communication, V2V communication always operates in a decentralized way, which causes more chaos under recent management technology. Therefore, it should only serve as an auxiliary method for the V2I communication system. For the V2I communication method, the ground-aerial vehicles commute to the stations in a centralized way, and the stations broadcast the ordered signals to the vehicle in its own communication zone. In our work, we mainly focus on the V2I communication pattern, which has more practical meaning than the V2V communication pattern. In the V2I communication, the amphibious ground-aerial vehicles send their planning airline, similar to the corridors for UAM, to the station. Then, the station communicates with other felicities to ensure the feasibility of its airline against other existing airlines or barriers like the skyscrapers in the urban area and sends the permission back to the vehicles\cite{zeng2021performance}. 

\par Recently, the roadway vehicular network technology is developed to thrive. For instance, the dedicated short-range communication (DSRC) technology varies from 75MHz to 5.9GHz band\cite{kenney2011dedicated}. The long-term evolution-vehicle (LTE-V) cellular technology which have large communication ranges, low latency, and frequent data exchange rates \cite{ahmad2019lte}. To meet the increasing bandwidth demand, the dynamic spectrum access (DSA) was proposed \cite{federal2010second}. Those roadway connected systems could fulfill the corresponding communication task.

\par
\begin{figure}
\hspace{-0.1cm}
\includegraphics[width=9cm]{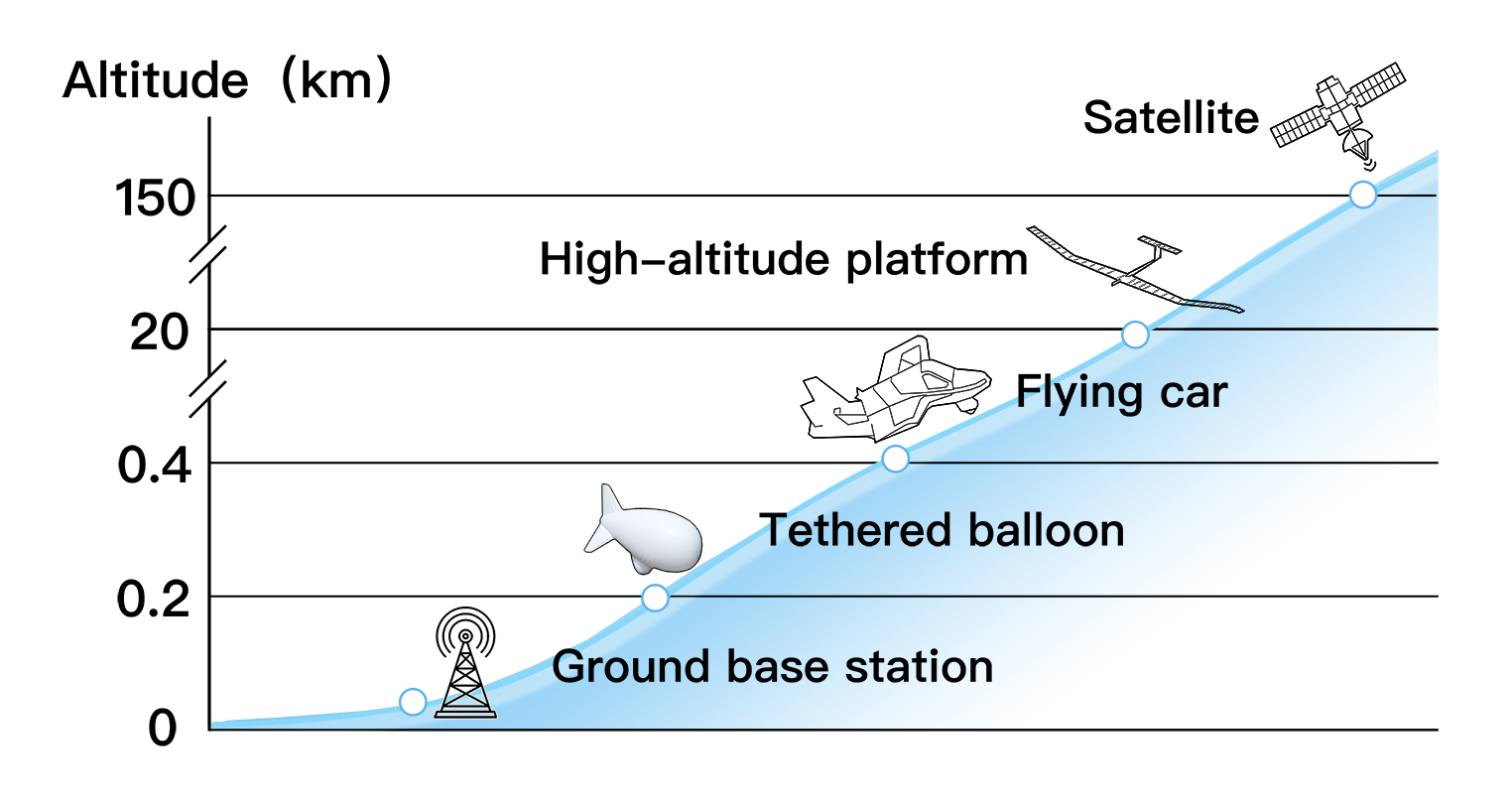}
\caption{ \centering Several potential communication stations for urban ground-aerial vehicles}
\label{communication}
\end{figure}

 Considering the requirement in efficiency and effectiveness, the existing on-ground communication technology could barely meet the demands for the flyable vehicles \cite{azari2019cellular}. Thus the amphibious vehicles would also suffer from the worse communication. Current roadway communication station concentrate its signals near ground but cannot cover the vehicle in almost 300-meter height, shown in Fig.\ref{communication}. Moreover, most communication stations focus their technologies on 4G radio. For those ground-aerial vehicles, which are in a consistently swift-changing state and demand extreme real-time signals, 4G radio communication shows weak performance in terms of high data delay and low connectivity. Hence, it is significant to involve new communication technology such as the Little Earth Orbit (LEO) satellites into 3D transportation system. 

\par Regarding the data delay and low connection rate generated by the flying altitude of the ground-aerial vehicle, one of the potential solutions is to place the base station (BS) not only on the ground but also in the air. Those aerial base stations could be tethered balloons (TBs), high-altitude platforms (HAPs), or satellites\cite{saeed2021wireless}. The tethered balloons belong to one of the low altitude platform family which always operates between 200 to 400 meters. They could take over the tasks for the telecommunications and broadband services. Compared to the GBS, TB has the advantages in the larger converge area, lower propagation delay, and higher operation altitude \cite{alsamhi2019tethered}. Regarding the high-altitude platforms (HAPs), they always install the buoyancy to float in the stratosphere at altitudes of up to 22 km, which is much higher than the TBs\cite{tozer2001high}. The higher operation height provides a larger coverage area, but still meets the requirement for communicating with air-driving vehicles. As for the satellite networks, they have the largest coverage area, almost a global coverage area, but suffer from the highest propagation delay compared to previous communication facilities. Constrained by their performance, they could operate as an auxiliary connected system which transfers information such as speed, altitude, location or the already planned trajectory between flying vehicles and the centralized control station.

\subsection{Near-ground Perception}

\par Compared with traditional autonomous driving in which altitude information is generally not required, the ground-aerial perception is more difficult. Sensing includes two tasks: internal and external. Internal perception mainly focuses on the vehicle itself and collects primary motion data such as location, velocity, altitude, and attitude. Its main purpose is to control flight stability. The external perception system aims to gain awareness of the circumstances. 

\par The perception hardware system is mainly composed of GPS, IMU, barometer, and light-stream module for internal sensing. One of the shortcomings of GPS originates from instability and precision floating due to the insufficient number of available satellites, especially in urban areas with skyscrapers blocking the signal transmission. Another drawback is that the inertial navigation system gains accumulated errors without short-term calibration. In an environment where the GPS is invalid, the optical flow module assists the position estimation of the vehicle. It dynamically evaluates the altitude by detecting the changes in pixels to determine the correspondence between adjacent frames. For the external perception, the meteorological issues cannot be neglected during flights, either. Because of the characteristic in short wave and strong penetration, radar demonstrate strong robustness to rainy or snowy weather and immunity to the darkness which blinds the camera. Miniaturization and low-cost meteorological RaDAR, one kind of millimeter-wave RaDAR, plays a key role in automatic flights to detect low-altitude turbulence and wind shear during a near-ground flight in complex environments, and to eliminate the negative effect on riding experience. \cite{yangjun2020progress}. In reality, these sensors including the millimeter-wave RaDAR, LiDAR, and vision sensors \cite{lu2018survey,araar2014new,cesetti2010vision} \textit{et al.} generally fuse into multi-sensor system to obtain more accurate data than the tedious perception system. Multi-modal data from sensors have the great potential to generate high standard results and take over more complex tasks\cite{carrillo2012combining,cesetti2010vision,ho2018adaptive}. In the following part, we organize three parts as follow: the perception assisted landing that aims to determine the landing site by visual guidance, perception for the non-cooperative objects when the amphibious vehicle works in the unknown or unstructured environment, and the datasets for training the near-ground perception.

\subsubsection{Perception Assisted Landing}
\par Take-off and landing are high-risk processes for flyable vehicles and some studies have illustrated that computer vision techniques can be applied to assist landing tasks. Region segmentation technology performs the satisfying results on vision-based landing. It can distinguish exact ground types, such as residential areas, rural highways, woods, and grasslands, to select the best landing sites. Segmentation problems are commonly summarized as pixel-level classification issues, which can be categorized into two classes: semantic and instance segmentation \cite{minaee2021image}. Semantic segmentation recognizes pixels belonging to one cluster but cannot recognize which cluster the objects belong to. Instance segmentation focuses on the recognition of target objects; however, is not suitable for the classification of overlapping objects \cite{osco2021review}. Based on previous studies, Kirillov \cite{kirillov2019panoptic} proposed a panoptic segmentation method to unify semantic and instance segmentation. In this work, the panoptic segmentation method takes the pixel-level annotation figure as input and trains the network to recognize near-ground lane lines. Most common aerial semantic segmentation studies capture the scene in the top view. However, compared to classic aerial photography, the scale of the scene captured in the oblique view is significantly distorted. To address this, \cite{Lyu2021} introduced a multi-scale attention network and fused the features captured from multiple scales to digest sufficient features. 

\par
\begin{figure}
\centering
\includegraphics[width=9cm]{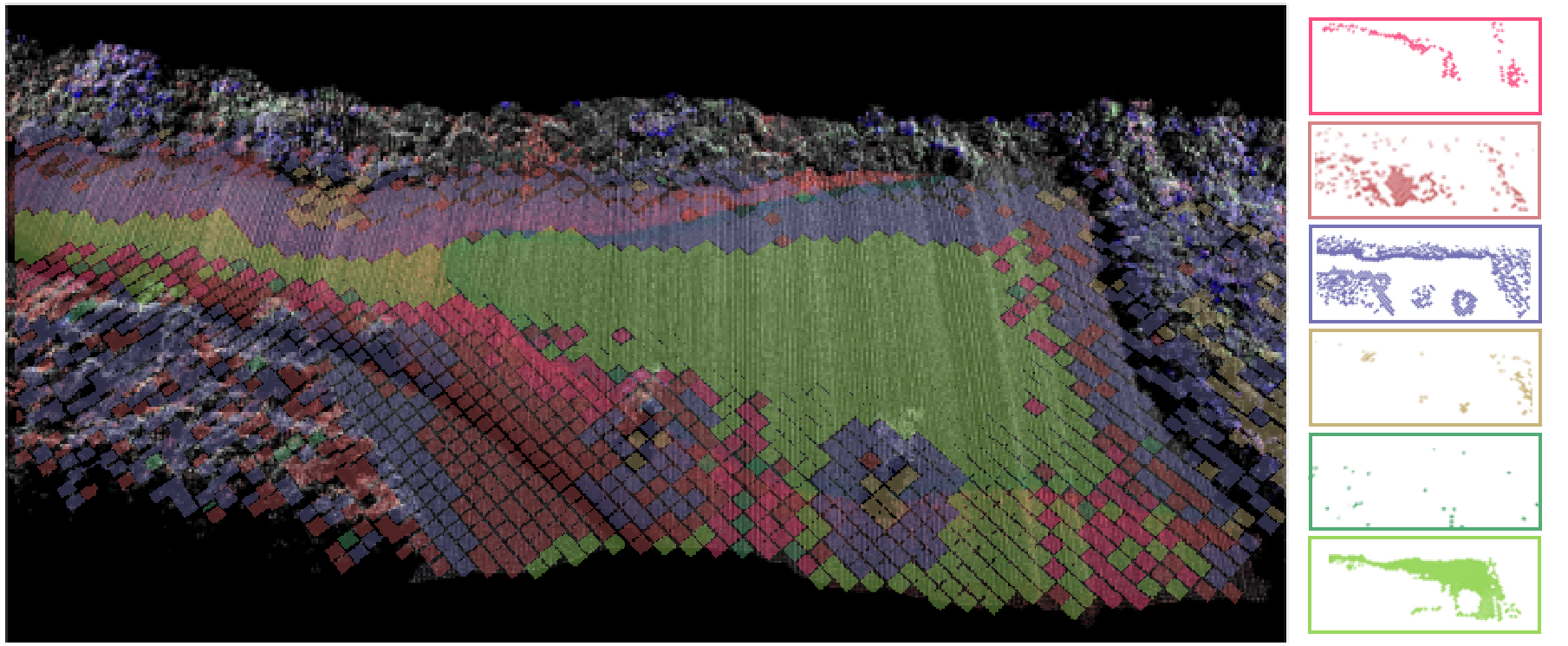}
\caption{ \centering
Selection of the landing zone. The figure on the top shows the results of the terrain slope. Six colors represent different grids: slope, slope and roughness, roughness, large spread, large residual, and flat.}
\label{landing zone selection}
\end{figure}

 In traditional autonomous ground vehicles, LiDAR is used to acquire the point cloud of surroundings, draw a 3D point cloud map and automatically identify the unknown terrain. Similar technologies can also be applied to the perception of amphibious vehicles. Using LiDAR sensors, \cite{Wallace2012, Khan2017, Fuad2018, Lin2019} separately displayed the near-ground perception of the forest, mountain, and seashore areas and built the corresponding digital terrain models. Laser generated by LiDAR has strong penetration, such that obtain information beneath shelters like trees to easily handle the three-dimensional segmentation task \cite{yan2018automated}. Furthermore,\cite{Hayton2020} analyzed the point cloud returned by a 3D LiDAR sensor with a convolutional neural network and detected humans walking on the ground. There was also a precedent for the research on airborne LiDAR-guided aircraft landing. As shown in Fig.\ref{landing zone selection}, \cite{Scherer2012} collected the real-time point cloud map to estimate the terrain based on a downward scanning LiDAR and evaluated the slope of the ground terrain through plane fitting methods. With these collected data, the evaluation towards the slope of the ground terrain is accomplished to perform segmentation and determines a feasible landing position.

\subsubsection{Awareness to Non-cooperative Objects}
\par While flying in an unknown environment, cooperative objects like other flying vehicles and landmarks can be avoid depending on the intelligent connected system. However, non-cooperative objects like bird flocks and unregulated vehicles should be detected by self-awareness ability. To avoid these kinds of dynamic obstacles in real time, target detection and tracking should have an efficient architecture to be arranged on board with low power computation and less memory consumption while maintaining the real-time performance and high accuracy\cite{Cazzato2020,Jin2019,de2015board}. For example,\cite{DeSmedt2015} tracked pedestrians using an on-board real-time tracking system from a near-ground top view. In addition, the corresponding methods should be adaptable to various weather and environmental light conditions because sunlight may be blocked by high buildings. To address this, researchers\cite{fu2020dr} proposed a tracking method based on a discriminant correlation filter, which has the function of illumination adaptation and improves the anti-interference to complex meteorological illumination in the air. Some researchers\cite{li2022all} also invented a tracker based on the discriminant correlation filter, which has the function of illumination adaptation and enhances the anti-interference ability under the change of complex meteorological illumination conditions in the air.

\begin{table*}[]
\setlength{\abovecaptionskip}{0.1cm}
\caption{ A summary of aeronautical datasets used for sensing missions over the last five years}
\begin{tabular}{lllllllll}
\hline
Dataset name & Years & Tasks                                                                                 & Camera & LiDAR & Other sensors & Dataset type               & Links                                                                                                           & Sourses   \\ \hline
UAV123        & 2016  & Detection/Tracking                                                                    & P      & -     & -             & Low altitude dataset       & -                                                                                                               &  \cite{Mueller2016} \\
UAV20L        & 2016  & Detection/Tracking                                                                    & P      & -     & -             & Low altitude dataset       & -                                                                                                               &  \cite{Mueller2016} \\
UAV123@10FPS  & 2016  & Detection/Tracking                                                                    & P      & -     & -             & Low altitude dataset       & -                                                                                                               & \cite{Mueller2016} \\
DTB70         & 2017  & Detection/Tracking                                                                    & P      & -     & -             & Low altitude dataset       & \begin{tabular}[c]{@{}l@{}}https://github.com/flyers\\ /drone-tracking\end{tabular}                             & \cite{Li2017} \\
UAVDT         & 2018  & Detection/Tracking                                                                    & P      & -     & Height sensor & -                           & \begin{tabular}[c]{@{}l@{}}http://www.cs.columbia.e\\ du/$\sim$vondrick/vatic/\end{tabular}                     & \cite{yu2020unmanned} \\
Aeroscapes    & 2018  & Segmentation                                                                          & P      & -     & -             & Low altitude dataset       &  \begin{tabular}[c]{@{}l@{}}https://github.com/ishann\\ /aeroscapes\end{tabular}                                 & \cite{Nigam2018} \\
VisDrone2019  & 2019  & Detection/Tracking                                                                    & P      & -     & -             & -                           & \begin{tabular}[c]{@{}l@{}}https://aistudio.baidu.com\\ /aistudio/projectdetail\\ /882508?shared=1\end{tabular} &\cite{zhu2018vision} \\
Mid-Air       & 2019  & \begin{tabular}[c]{@{}l@{}}Real-time positioning\\  mapping/Segmentation\end{tabular} & P      & P     & GPS/IMU       & Low altitude dataset       & https://midair.ulg.ac.be/                                                                                       & \cite{Fonder2019} \\
UAVid         & 2020  & Segmentation                                                                          & P      & -     & -             & -                           & https://uavid.nl/                                                                                               & \cite{Lyu2020} \\
AU-AIR        & 2020  & Detection                                                                             & P      & -     & GPS/IMU       & Ultra-low-altitude dataset & \begin{tabular}[c]{@{}l@{}}https://github.com/bozcani\\ /auairdataset\end{tabular}                              & \cite{bozcan2020air} \\
SensatUrban   & 2020  & Segmentation                                                                          & P      & P     & RTK/GNSS      & -                           & \begin{tabular}[c]{@{}l@{}}http://point-cloud-\\ analysis.cs.ox.ac.uk/\end{tabular}                             &\cite{hu2021towards} \\
Campus3D      & 2020  & Segmentation                                                                          & P      & P     & GPS           & High altitude data set      & \begin{tabular}[c]{@{}l@{}}https://github.com\\ /shinke-li/Campus3D\end{tabular}                                & \cite{li2020campus3d} \\
Toronto-3D    & 2020  & Segmentation                                                                          & P      & P     & GNSS          & -                           & \begin{tabular}[c]{@{}l@{}}https://github.com\\ /WeikaiTan/Toronto-3D\end{tabular}                              & \cite{tan2020toronto} \\ \hline
 \setlength{\abovecaptionskip}{0.5cm}
\end{tabular}

	\label{tab2}
\end{table*}

\subsubsection{Dataset for Near-ground Perception}
\par We reviewed and compared the near-ground perception datasets in the recent five years, as shown in table.\ref{tab2}. Most of these datasets are collected using UAVs and can be used for near-ground perception, including target monitoring, tracking, and segmentation. Datasets are mainly generated by cameras, but part of the data is gathered using LiDAR with the assistant data from IMU and GPS. Photographs taken at very high altitudes have little value for near-ground perception. Therefore, after screening datasets from the aspects of size, task, scene, and the attitude of the UAV-collected data, UAV123, UAV20, LUAV123@10FPS\cite{mueller2016benchmark}, DTB70, and Aeroscapes\cite{bosch2019captioning} can meet the requirements to study near-ground perception.

\subsection{Decision-making and Planning}
\par Decision-making and path planning are imperative to achieve autonomous 3D transportation. The intelligence decision-making module should contain two functions: (1) vehicles should be aware of global traffic conditions and choose the best route; (2) vehicles should plan and replan trajectories based on local environmental awareness. This part mainly consists of ground-aerial navigation and decision-making technology, which includes path planning and motion mode selection, \textit{et al}. In addition, to ensure safety, the short-term recovery decision mechanism and intelligent emergency landing strategy should be considered.

\subsubsection{Ground/Aerial Navigation}
\par With the fusion of ground and air traffic data, the planning of an optimal ground/aerial path is needed but challenging. Ground-aerial navigation and path planning should generate a 3D route concerning various constraints\cite{wang2014study}:
\begin{itemize}
    \item Physical realizability: amphibious vehicles must obey the physical and dynamic limitations. Physical limitations may include the maximum cruise range, minimum land/taking-off distance, and energy cost either during flight or driving on the ground. Dynamic limitations consist of the maximum and minimum turning radius, steepness of climbing or descent angle, and the highest and the lowest flying altitude\cite{zhao2018survey}.
    \item The task requirement: the task requirement is allocated according to the task itself, for instance, the endurance and the fuel consumption.
    \item Real-time decision-making: 3D path planning must have a high computation efficiency. Furthermore, path re-planning is required and plays a significant role in the response to unpredictable threats\cite{zhao2018survey}.
\end{itemize}

\par Based on the methods of ground/air traffic flow convergence mentioned in section \uppercase\expandafter{\romannumeral2}, there are different route design methods. For UAM, the air routes are almost planned along a straight line to save time. It starts from the ground/aerial departure station directly towards the destination. Under the straight-line planning method, the selection of the landing location based on the topological structure becomes the key \cite{straubinger2020overview}.

\par
\begin{figure}
\centering
\includegraphics[width=8cm]{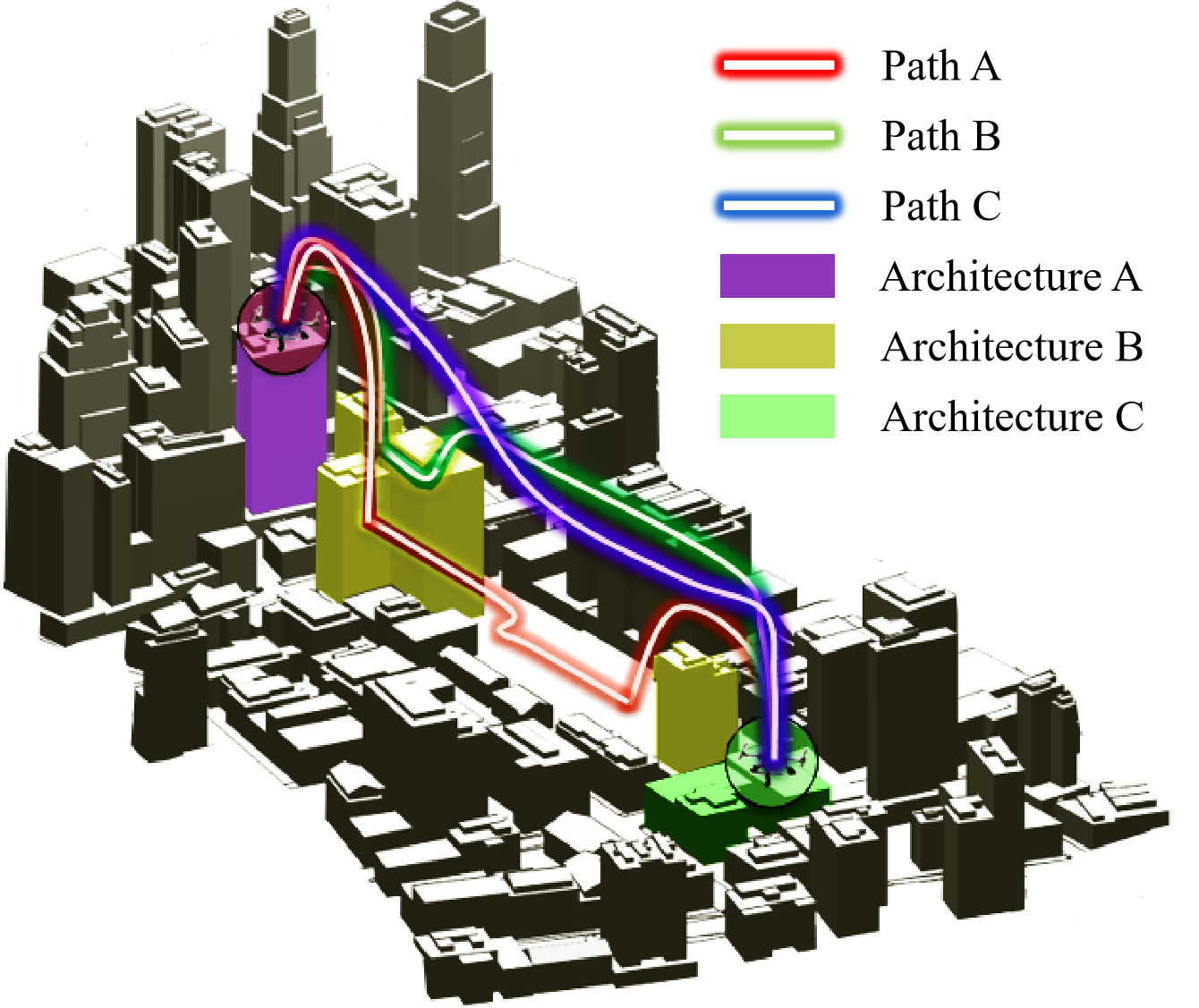}
\caption{ \centering
 The map search method results in a 3D navigation path planning for urban ground-aerial vehicles in the city. In Path A, the ground-aerial vehicle is supposed to both fly and drive. In Path B, the urban ground-aerial vehicle selects flying as the only locomotion mode; however, it lands on architecture B in half. In Path C, the urban ground-aerial vehicle determines to fly over the architecture with no pause.}
\label{planning in city}
\end{figure}

 Furthermore, because amphibious ground-aerial vehicles have multiple locomotion capabilities, navigation requires a 3D path planning and motion switching method. Presently, relevant studies on automatic ground/aerial path planning have mainly focused on three-dimensional global navigation and dynamic local online replanning. Most studies apply graph search and probabilistic algorithms \cite{sharif2019new}, such as the 3D rapidly exploring random trees or probabilistic road maps, and the algorithms in the 3D A* family. These methods generate an available path relying on the global information that is already known. Based on the principle of minimizing energy consumption, \cite{TerrySuh2020} proposed a state-space graph programming method based on the approximate dynamic programming theory. Researchers add up fuel consumption by different modes and mode-switching, and then set the consumption as the cost function for optimization. Finally, this provides a feasible strategy to guide the vehicle to cross the valley. Other studies \cite{sharif2018energy} applied three-dimensional A* algorithms in grid maps and took the terrain, obstacles, flying energy limit, and expected flight time into consideration. The 3D path planning for urban applications has been resolved by the graph planning method, as shown in Fig.\ref{planning in city} \cite{suh2020energy}. The graph-searching method has several limitations. First, navigation requires global information as a vital prerequisite. Second, global graphic path planning focuses on the macro map; however, details are neglected. One obvious negligence is that the length of the path is proportional to the energy consumed, which is unrealistic. Meanwhile, some trajectories are over-theoretical and ignore the kinematic characteristics. To fix this, \cite{choudhury2019dynamic} introduced a double graphic planning method and an inspector such that the over-idealized routes are forced to fit the kinetic features.

\par Instead of graphic path planning, paths could also be generated by intelligent algorithms such as particle swarm optimization (PSO)\cite{Zhang2013}, ant colony optimization\cite{Liu2017}, grey wolf optimization algorithm, simulated annealing algorithm, and genetic algorithm \cite{Sahingoz2013}. Intelligent computation inspired by nature could provide an outstanding solution to the nonlinear problem, which is challenging to model with conventional mathematics. For example, \cite{Ran2011} plans the path with PSO and sets the influence of path length as well as environmental medium of motion as the fitness function.

\subsubsection{Dynamic Path Planning}
\par Dynamic path replanning is another key component for the intelligent near-ground flight. Replanning should be done in real time to adapt to a fast-changing environment and determine the available routes \cite{subosits2019racetrack}. In addition, rapid path replanning is an emergency response operation that could resolve a crisis when the amphibious vehicle faces other vehicles and dynamic obstacles or is supposed to land immediately. Some methods based on the probability map constructed from perceptual data provide solutions for the problems of safety area assessment \cite{Loureiro2020}. Moreover, the model-free deep reinforcement learning method used in dynamic path planning performs well in obstacle avoidance. As an end-to-end controller, DRL incomes data such as the depth image, point cloud data captured by LiDAR, elevation map, and obstacle topological map, and outcomes of the policy that directly controls the motion, which may have a bright future for end-to-end path planning \cite{josef2020deep,yan2020towards,theile2020uav}. However, although learning-based approaches have behaved extraordinarily in simulations, the simulation-to-reality gap still confuses these approaches and blocks them from real-world use. Methods used in global path planning could theoretically also be used in this mission.

\subsubsection{Emergency Decision Support System}
\par To ensure the safety of passengers, vehicles, and pedestrians on the ground, emergency operation is indispensable \cite{hubmann2018automated}. The emergency operation system executes a contingency plan when a flyable vehicle is exposed to safety hazards and takes over the driving system to perform the recovery mechanism or landing \cite{di2017evaluating}.

\par The decision support system should be preconceived. Before take-off, several landing sites should already be determined. In addition, an online replanning system should be designed to offer online predictive redundant sites for landing. It offers the driver a list that includes all alternative emergency landing sites\cite{ayhan2018semi}. Considering the essential attributes such as accessibility of crash landing location and distance from dense population, \cite{Coombes2016} used the Bayesian decision network to select the best-forced landing location from a plethora of available candidate landing sites.

\par Meanwhile, the air emergency decision-making needs to respond to the volatile near-ground environment timely and quickly. Real-time decision-making is based on a probability map generated by the perception system, including the RaDAR, LiDAR, and vision sensors\cite{loureiro2020survey}. The probability map could offer adequate information and lead to a marked improvement in the emergency landing system. According to the upper-layer perception information, emergency algorithms are used to reduce damage to passengers. Researchers have proposed an architecture of Safe2Ditch that avoids ground obstacles in emergencies and plans landing routes based on UAV platforms \cite{lusk2018vision}.\cite{Arantes2018} solved the problem of path replanning to improve autonomous response capabilities in critical situations. Furthermore, damage to the ground needs to be reduced to a minimum. For instance, \cite{gonzalez2021visual} proposed a safe landing area tracking algorithm based on Kalman filtering to guide emergency landing and prevent ground personnel from damage.

\subsection{Intelligent Control Technologies} 
\par As the core of the flying, low-level control determines the performance of an amphibious vehicle. While designing a fly driving control system, control allocations vary among different configurations. Researchers have studied the control rate of vehicles in different configurations based on Lyapunov's stability theory. Presently, most of the flight modes of land and air vehicles draw lessons from multi-rotor step proportional-integral-derivative (PID) series execution and control methods \cite{Mintchev2018,Meiri2019}. Moreover, approaches such as optimal control, robust control, backstepping control, adaptive control, fuzzy control, and even learning-based methods have been applied to flight and driving control systems \cite{kim2013study,paden2016survey,jiang2018lateral}. Currently, the main difficulties in intelligent control include dynamic fast switching between modes and ground effect compensation for near-ground flights.

\subsubsection{Fast Mode Switching Control}
\par Fast modes changing means the process of ground-aerial transition. Ideally, the actuators should work cooperatively to ensure a smooth handover of the land-air mode switching. The detailed dynamic model was studied in \cite{tan2021multimodal}, where a comprehensive dynamic model is proposed to describe continuous air and ground motions. In this study, a model-based predictive control method and two-stage control distributors effectively improve the landing stability of vehicles. In view of the air-land switching control, the mode reference adaptive system is added to the linear quadratic regulator algorithm to solve the problem of uncertain model parameters during the switching process. Besides, researchers designed a fuzzy PID controller to control the ground movement of the vehicle, which can quickly accomplish switching when encountering obstacles. In addition, researchers from Tsinghua University proposed a comprehensive dynamic model to describe the continuous aerial/ground motion and adopted the model predictive control method and the design of a two-stage control distributor to effectively improve the landing stability of flying vehicles \cite{tan2021multimodal}.

\subsubsection{Offset Issues by Ground Effect of Near-ground Flight}
\par Amphibious vehicles with rotary wings suffer from severe ground effects. The stability of take-off and landing is always a barrier for flight control systems \cite{keshavarzian2020pso}. When the ground effect occurs, the airflow obstructed by the ground generates aerodynamic interference and influences the normal airflow\cite{matus2021ground}. Modeling the near-ground air fluid is a complex task because the arguments in the model are massive and suffer from the curse of dimensionality\cite{griffiths2002study,curtiss1984rotor,krishnan2016numerical}. Although modeling the ground effect is intricate, the offset to the ground-aerial vehicle is still worthy of discovery. With the theoretical analysis of the ground effect on propellers,\cite{cheeseman1955effect} derived the ground effect function with forward velocity as an independent variable. In addition, \cite{Bernard2017} built a ground effect measurement platform specially used for multi-rotor aircraft and studied the control systems in bounded disturbances by system identification. Some studies also build a ground effect test platform that can accurately obtain observation results and improve near-ground flight stability\cite{Conyers2018}.

\par To address the above problems, compensation controllers are needed. For instance, \cite{Keshavarzian2020} invented a backstepping controller that could optimize the parameters from the controller using a PSO algorithm. The key dominance of the controller is its real-time estimation and compensation of the ground effect dynamically. He also introduced a nonlinear state-space model to eliminate the ground effect in another study \cite{Keshavarzian2019}, and measured the ground effect with the surface coefficient \cite{Sanchez-Cuevas2017}. Regarding to the non-linear system, the neuroadaptive control provides a brand new solution for the disturbance compensation and rejection \cite{yang2022neuroadaptive, yang2022neuroadaptive1}. The local ground effect phenomenon of a multi-rotor is proposed for the first time. The experiment shows that the aircraft perfectly leaps over obstacles and tracks the prescribed path at a ground height of twice or thrice the propeller length.

\section{Technical Barriers and Potential Solutions}

\par The existing amphibious vehicles prove the technical feasibility, but there still exist huge gap in forming into the transportation system. Although intelligent technologies can be deployed on amphibious vehicles and effectively improve the autonomy of transportation, many crucial points still need further development. Previous sections already discuss the advantages of evolving the intelligence into 3D transportation system and vehicles. In this section, we summarize the core technical barriers in the aspect of both individual and swarm, and discuss the corresponding potential solutions.

\subsection{Modality Switch}

\par Presently, terrestrial driving and aerial piloting systems are not systematically integrated. Although amphibious vehicles like Aeromobil redesigned the rudder for both flying and driving, the transition between multi-modal locomotion cannot be performed smoothly and continuously yet, and relevant research works still stay at the developing stage.

\par The switching between two modes is regarded as the biggest barrier that limits the flexibility of vehicles. Most current flying cars require space and speed for landing and take-off, such as the hybrid car-aircraft AirCar by Klein Vision requires 350m length runway and 115km/h to take off. Moreover, the vehicle must be stationary during mode switching. For example, all amphibious vehicles shown in Fig.\ref{klein vision} to Fig.\ref{PAL-V} require to stay static when transit between different modes, and this process also takes time. Static switching leads to the discontinuity in movement which undoubtedly puts the amphibious vehicles in an awkward and passive position. Those two disadvantages greatly restrict the mobility of amphibious vehicles and impedes its integration into urban transportation. There are several potential solutions by using the current techniques to tackle this dilemma.

\par The short take-off and landing (STOL) could serve as one of the solutions. STOL is widely studied by aircraft designer to reduce the runway length, especially for small airplanes due to general aviation airports usually have limited area. This technology can effectively reduce the runway length and may be potential for current existing amphibious flying cars. \cite{courtin2018feasibility} uses the Geometric Programming to optimize the structure of fixed wing UAM and already proves that the runway length could reduce into 30-100 metres. Under the particular structure, the speed for take-off and landing decrease sharply. However, the established model is very theoretical, and no real-world experiments are generated. In addition, the number of onboard passengers is quite low which equals to 4-5. The efficiency is the most serve problems it meets in the future.

\par Amphibious vehicles with rotary wings show more potential for seamless and dynamic switching. Researchers have worked out the dynamic model and produce the model predictive control (MPC) based rotor controller to compensate for disturbance{\cite{tan2021multimodal}}. The landing trajectory is planned smoothly by the control theory and model proposed in the article which shows great significance for the future manned amphibious vehicles.

\par Besides, thrust vectoring technology is a prospective method for super-maneuver flight control. For instance, tilt-rotor techniques can empower engineers in areas of development that might be considered outside of their traditional remit. The propulsion of vehicles equipped with this technology is capable to change the direction to better meet the complicated and rapidly varying balanced condition. For example, the BlackFly \cite{openerOpeneraero}, which is a single seat flyable vehicle designed by Opener, can take-off and land vertically in a narrow room by changing the propeller direction. In fact, tilt-rotor technology has been widely applied to the emerging eVTOL UAM vehicles and its feasibility has been already tested. Thus, employing tilt-rotor on amphibious vehicles is rationally potential to make ground-air transition more smooth and comfortable.

\subsection{High-speed Perception}

\par Recent perception technologies are incapable of taking over the sensing task for amphibious vehicles due to the high-speed moving. For the present on-ground autonomous driving sensing tasks, cameras and LiDAR play the dominant role. However, accuracy by camera is suffered from the dynamic motion, and LiDAR also shows its weakness in high-speed environments. Generally speaking, the top limit speed of autonomous driving is around 130 km/h. For instance, regular Tesla autopilot Autosteer is limited to a maximum speed of about 135 km/h\cite{teslaTransitioningTesla}. While amphibious vehicles like Aeromobil 4.0 can fly at a speed of 360 km/h, and the PAL-V also reaches its maximum speed of 180 km/h. Thus, the near-ground perception for fly-driving requires more sensitive and accurate devices to fit the high-speed environment. For now, there are indeed some perception application experiments under high-speed conditions. For example, the world's maximum speed by fully autonomous driving was created by team PoliMOVE from Italy’s Politecnico di Milano and the University of Alabama using a self-driving race car, in 2022. An astonishing world record speed of approximately 309.3 km/h was reached \cite{newatlasAutonomousRace}. Nevertheless, this test was achieved on a straight runway without any obstacles or uncertainties. 

\par As for the dynamic blur problem for LiDAR, the sensor is constrained by the upper limit number of the laser beam and the point clouds contain relatively thinner information than images \cite{kabzan2020amz}. Thus, motion deblurring methods for the LiDAR are struggled to achieve. In contrast, the camera is less constrained by hardware and contains information that is rich enough to reconstruct\cite{kabzan2019learning,weiss2020deepracing}. In the following paragraphs, we only focus on the motion deblur for images, which is much easier to access and contains a greater probability of success.

\par The key insight of motion deblurring for images depends on how to improve the computation speed and the algorithms for motion deblurring. The former one can be solved by deploying a high-performance computer and lightweight algorithms onboard. Edge computing and distributed computing could also solve this. By contrast, motion blur is more challenging. The motion blur occurs because of the relative placement between camera and target\cite{sada2018image}, which could be caused by bumps during landing or high-speed near-ground cruising. Since similar problem happens in ground autonomous driving, it has drawn sufficient attention and several existed solutions could be derived\cite{nah2021ntire}.

\par Image blur could be solved by image processing. Based on the Convolutional Neural Network (CNN), Schuler \textit{et al}. \cite{schuler2016learning} proposed a architecture with CNN for kernel estimation. The network is composed of several groups of blocks stacked in sequence, and each block includes feature extraction, kernel estimation, and image estimation. Feature extraction module extracts feature from the blurry figure and sends into kernel estimation module that approximated the uniform blurry kernel. The latent image which equals to the predictive clear image is generated by image estimation. As the early CNN-based figure deblurring algorithm, it has certain advantages compared with the algorithms of the same period. However, its shortcoming is undeniable that the kernel shape is fixed and inability to deal with irregularities or large blurs. For the non-uniform blur, scene depth \cite{hu2014joint,paramanand2013non} or segmentation \cite{hyun2013dynamic} is always selected to assist.

\par With the development of figure deblurring, some algorithms eliminate the kernel estimation. S Nah \textit{et al}. \cite{nah2017deep} adopt a multi-scale network architecture for image deblurring and choose to record the sharp information at different scales. The lower scale image goes through up-convolutional process and forms into the higher scale. The final clear image is obtained after three-time accumulation. The kernel-free method avoids the kernel estimation process and gains higher complexity. Meanwhile, as the step decreases, the computation efficiency is promoted obviously\cite{noroozi2017motion}. However, the pure kernel-free method without a particular design cannot generate sufficient details and hardly improve the spatial resolution\cite{zhang2018gated}.

\par Moreover, motion blur often mixes with the problem of low-resolution quality. To address this, \cite{zhang2018gated} present a dual-branch architecture, named Gate Fusion Network (GFN) including motion deblurring module, super-resolution feature extraction module and a gate module. The first two modules each occupy one branch to solve the target problem and the gate module tries to adaptively fuse the features of deblurring and super-resolution branches. The performance of GFN is beyond reproach, but its low computational efficiency is the biggest weakness.

\par Compared with other networks, generative adversarial networks (GAN) owns the great potential towards this problem. In 2014, Goodfellow \textit{et al.} invented GAN in generating face images and the hand-written numbers\cite{goodfellow2014generative}. Over the past few years, GAN was quickly applied into the field of image deblurring with its powerful detail generation skill. Wei \textit{et al}. attempted GAN network for motion deblurring, and extended scene into the rainy deblurring by transfer learning\cite{wei2021dynamic}. As the losses in his work, the Wasserstein distance \cite{arjovsky2017wasserstein} was selected for the discriminator and Perceptual loss \cite{johnson2016perceptual} as the loss function of generator. Several articles \cite{ledig2017photo,kupyn2018deblurgan} proved that it not only have a strong effect on motion deblurring, but also could deal with details processing and high-resolution or even super-resolution image prediction.

\subsection{Adaptive Landing}

\par Generally speaking, landing is the most dangerous part of flight\cite{alam2021survey}. Amphibious vehicles face more difficulties compared with traditional aircraft. Above all, amphibious vehicles' landing trajectories are more complicated than an aeroplane that uses fixed landing sites. For example, the trajectory in UAM transportation system is a vertically straight course. Meanwhile, landing in urban areas means complex environment, in which the on-ground traffic participants, such as pedestrians, cyclists, and vehicles, are ever-changing and hard to forecast. Moreover, vehicles are supposed to remain a high speed after the touchdown point, which adds more burden to generate a smooth landing trajectory. Furthermore, there is no guidance that can be followed and the landing site has to be generated according to the rapidly changing circumstances. These challenges raise the adaptive landing of amphibious vehicles to an even higher level of difficulty than traditional flyable vehicles.

\par For the adaptive landing of an amphibious vehicle, the landing trajectory must be determined in advance for better preparation. Except for the static obstacles like buildings which are efficiently dealt with by advanced path planning, the moving obstacles like flying birds and surrounding vehicles also hinder to land. Dynamic path planning for obstacle avoidance is also required. It demands a high standard for calculation time to generate avoidance trajectory and the avoidance distance.

\par Studies on trajectory planning with constrains could help to solve landing with speed. In order to ensure that the amphibious vehicle lands at a particular designed place with a demanded speed, \cite{haghighi2022performance} planned the trajectory with Dubins Curve for fixed-wing aircraft and optimized it by the meta-heuristic structure. It adapted to multiple middle turns to avoid the obstacles on the route. In the process of trajectory generating, the environmental effects such as wind and the dynamic module served as the constrained function. In terms of simulation, this work chose Airbus A320 as the aircraft in the experiment. Although it is different from our amphibious vehicle in the dynamic model, the path planning method has great potential to be derived. The main disadvantage lies in the computational time, which is high due to the meta-heuristic optimization and becomes a burden to the on-board computation unit.

\par Except for global trajectory generation, dynamic path planning should also be concerned. Lin \cite{lin2020fast} proposed a fast geometric avoidance (FGA) algorithm for fixed-wing aeroplanes. It classifies the obstacles, determines the probable conflicts by spatial location, relative velocities, angular displacements, as well as collision cone concepts, and calculates the safe distance \cite{goss2004aircraft,daniels2012collision}. The algorithm searches for the next attainable way-point along the original path after the aircraft has manoeuvred to a safe distance, and the vehicle quickly recovers to the originally intended path. Compared to the way-point generation or optimal path planning approaches, which often recalculate whole path lengths, FGA decreases the maneuvering distance to avoid obstacles, and also speeds up calculation time. However, the velocity for aircraft in assumption ranges from 13 m/s to 19 m/s, which is lower than our predetermined speed but still meaningful to refer to.

\par Beside the dynamic obstacle avoidance, the adaptive landing is also required to reconstruct the unknown environment and identifies the potential landing site. \cite{iiyama2021deep} selected the deep reinforcement learning method to control the aircraft to land in an unknown environment. The observation module captures figures for the landing site and reconstructs the landing site. Because of its online planning method, the aircraft can still land safely even if the environment is changing. As for the landing of amphibious vehicles, this work is instructive and could be directly derived if the observation method switches into near-ground perception such as stereo ranging\cite{theodore2006flight}, optical flow\cite{cesetti2010vision}, and color segmentation\cite{edwards2007vision}.

\subsection{3D Traffic Flow Control and Driving Safety}

\par A mature 3D transportation system must introduce more amphibious vehicles to airspace \cite{Yang2020} and should endurance thousands of flights in the metropolitan areas every second. What followed is the 3D traffic flow control becomes a severe problem \cite{Siewert2019} since flyable vehicles entail higher degrees of efficiency and safety \cite{ellis2020time}.  Multiple ground-aerial vehicles require a high level of real-time monitoring to conduct continuous trajectory management to ensure safety\cite{Bosson2018}. 

\par The amphibious vehicle is unrestricted by the roads and has access to generate the path in 3D space freely. It turns our problem closer to aircraft management than traditional on-ground traffic flow control. In traditional airflow control, real-time air traffic management (ATM) is crucial\cite{mathur2019paths} and its safety is achieved in the human-within-the-loop pattern, for instance, the Air Traffic Controller (ATCO)\cite{degas2022survey}. Obviously, with the expansion of the 3D traffic system, management systems with a large proportion of human participants are not well-suited to address the management problem in real-time communication, surveillance and airline planning for amphibious vehicles. For traffic flow control, human-involved control includes subjective considerations unavoidably and is hard to analyze the global situation comprehensively. Regarding driving safety, the size of overall traffic information which contains the disaster amount of data is tremendous for human to manage. Any tiny bias of the control system could cause a series of tragedies. To sum up, the intelligent solution is crucial.

\par To address the complex management problem, we propose the potential solutions from two perspectives: the global perspective and the local perspective. From the global perspective, we pay more attention to the workload balance. The workload balance methodology is committed to reduce the workload differences between sub-regions and seek the workload allocation approach that gains the most globally efficient\cite{chen2018workload}. From the local perspective, we try model-free control methodology to address the current problem in distributing the swarm of vehicles without collision and to maintain the driving safety for vehicles.

\par In the global perspective of traffic management, the workload balance would be accomplished by decomposing the whole region into several sub-regions and narrowing the workload gap among them. For traffic control, it reduces the probability of local congestion. Turning to the communication systems, the oversized local traffic flow will add too much burden to the communication station and result in communication delays, which is fatal for amphibious vehicles. As for the tool to decompose region, the Voronoi Diagram \cite{boots2009spatial} is a conventional option \cite{wu2011distributed}. Xue \cite{xue2009airspace} proposed a methodology to distribute the aircraft in certain airspace and optimize the Voronoi pattern with Genetic Algorithms (GA). Firstly, the region is decomposed randomly and the algorithm evaluates the current cost value. Then the pattern is optimized by GA and the cost value is recalculated. The decomposition repeats until the terminal condition is satisfied. After the precise segregation, the workload is distributed evenly and the simulation shows that the average flight time reduces.

\par In the local management control, safety management becomes the major point. More precisely, the unsafe factor mainly comes from collisions between adjacent vehicles. To address the mentioned issue, decentralized multi-agent path planning offers lessons to avoid collision among vehicles and guarantee safety. PRIMAL$_{2}$ \cite{damani2021primal} is a decentralized multi-agent path-generating framework with reinforcement learning and imitation learning. Each agent takes partial observation and plans the available path without any collision. The multi-agent structure is made up of a threefold approach. Firstly, the harmony and rational movements in a particular situation concerning the surroundings are identified. Then, the forecast about the future states, actually obtained by broadcasting to the surrounding agents and receiving the already planned movements, is determined. Finally, the favorable policy is learned through the centralized planner using imitation learning, and migrated to individuals to instill mature policy. Compared to its previous work PRIMAL\cite{sartoretti2019primal}, it significantly scales up the number of agents to 2048. With respect to the distribution of amphibious vehicles, PRIMAL$_{2}$ shows great potential and is extremely efficient in collision avoidance between vehicles. However, PRIMAL$_{2}$ simplified the working condition into a two-dimensional scene which assumed the agents are on the same altitude. There is still a considerable gap between 2D and 3D multi-agent path planning, and requests for deeper studies. 

\section{Conclusion}

\par We elaborate the background of amphibious vehicle, and select four representative examples to extract their features and distinguish the difference against UAM. We then summarize the intelligent technologies such as the connected systems, near-ground perception, decision-making \& path planning, and modal-switching control, for the future 3D transportation which summons numerous amphibious vehicles and supportive facilities. On the basis of current technology, the barriers encountered in future 3D traffic are presented as modality switch, high-speed perception, adaptive landing and traffic management, and we offer the potential solutions correspondingly.

\section*{Acknowledgment}
This work was supported by the National High Technology Research and Development Program of China under Grant No. 2018YFE0204300, and the National Natural Science Foundation of China under Grant No. U1964203, and sponsored by Meituan and Tsinghua University-Didi Joint Research Center for Future Mobility.



%

\bibliographystyle{IEEEtran}
\bibliography{mybibliography.bib, add.bib}

\begin{thebibliography}{100}
\providecommand{\url}[1]{#1}
\csname url@samestyle\endcsname
\providecommand{\newblock}{\relax}
\providecommand{\bibinfo}[2]{#2}
\providecommand{\BIBentrySTDinterwordspacing}{\spaceskip=0pt\relax}
\providecommand{\BIBentryALTinterwordstretchfactor}{4}
\providecommand{\BIBentryALTinterwordspacing}{\spaceskip=\fontdimen2\font plus
\BIBentryALTinterwordstretchfactor\fontdimen3\font minus
  \fontdimen4\font\relax}
\providecommand{\BIBforeignlanguage}[2]{{%
\expandafter\ifx\csname l@#1\endcsname\relax
\typeout{** WARNING: IEEEtran.bst: No hyphenation pattern has been}%
\typeout{** loaded for the language `#1'. Using the pattern for}%
\typeout{** the default language instead.}%
\else
\language=\csname l@#1\endcsname
\fi
#2}}
\providecommand{\BIBdecl}{\relax}
\BIBdecl

\bibitem{fan2019autonomous}
D.~D. Fan, R.~Thakker, T.~Bartlett, M.~B. Miled, L.~Kim, E.~Theodorou, and
  A.-a. Agha-mohammadi, ``Autonomous hybrid ground/aerial mobility in unknown
  environments,'' in \emph{2019 IEEE/RSJ International Conference on
  Intelligent Robots and Systems (IROS)}.\hskip 1em plus 0.5em minus
  0.4em\relax IEEE, 2019, pp. 3070--3077.

\bibitem{qin2020hybrid}
Y.~Qin, Y.~Li, X.~Wei, and F.~Zhang, ``Hybrid aerial-ground locomotion with a
  single passive wheel,'' in \emph{2020 IEEE/RSJ International Conference on
  Intelligent Robots and Systems (IROS)}.\hskip 1em plus 0.5em minus
  0.4em\relax IEEE, 2020, pp. 1371--1376.

\bibitem{kalantari2014modeling}
A.~Kalantari and M.~Spenko, ``Modeling and performance assessment of the hytaq,
  a hybrid terrestrial/aerial quadrotor,'' \emph{IEEE Transactions on
  Robotics}, vol.~30, no.~5, pp. 1278--1285, 2014.

\bibitem{kalantari2013design}
------, ``Design and experimental validation of hytaq, a hybrid terrestrial and
  aerial quadrotor,'' in \emph{2013 IEEE International Conference on Robotics
  and Automation}.\hskip 1em plus 0.5em minus 0.4em\relax IEEE, 2013, pp.
  4445--4450.

\bibitem{itasse2011equilibrium}
M.~Itasse, J.-M. Moschetta, Y.~Ameho, and R.~Carr, ``Equilibrium transition
  study for a hybrid mav,'' \emph{International Journal of Micro Air Vehicles},
  vol.~3, no.~4, pp. 229--245, 2011.

\bibitem{elsamanty2012novel}
M.~Elsamanty, M.~Fanni, and A.~Ramadan, ``Novel hybrid ground/aerial autonomous
  robot,'' in \emph{2012 First International Conference on Innovative
  Engineering Systems}.\hskip 1em plus 0.5em minus 0.4em\relax IEEE, 2012, pp.
  103--108.

\bibitem{sarica2019technology}
S.~Sarica, B.~Song, J.~Luo, and K.~Wood, ``Technology knowledge graph for
  design exploration: application to designing the future of flying cars,'' in
  \emph{International Design Engineering Technical Conferences and Computers
  and Information in Engineering Conference}, vol. 59179.\hskip 1em plus 0.5em
  minus 0.4em\relax American Society of Mechanical Engineers, 2019, p.
  V001T02A028.

\bibitem{sarica2021idea}
S.~Sarica, B.~Song, J.~Luo, and K.~L. Wood, ``Idea generation with technology
  semantic network,'' \emph{AI EDAM}, vol.~35, no.~3, pp. 265--283, 2021.

\bibitem{stoeter2002autonomous}
S.~A. Stoeter, P.~E. Rybski, M.~Gini, and N.~Papanikolopoulos, ``Autonomous
  stair-hopping with scout robots,'' in \emph{IEEE/RSJ international conference
  on intelligent robots and systems}, vol.~1.\hskip 1em plus 0.5em minus
  0.4em\relax IEEE, 2002, pp. 721--726.

\bibitem{lambrecht2005small}
B.~G. Lambrecht, A.~D. Horchler, and R.~D. Quinn, ``A small, insect-inspired
  robot that runs and jumps,'' in \emph{Proceedings of the 2005 IEEE
  international conference on robotics and automation}.\hskip 1em plus 0.5em
  minus 0.4em\relax IEEE, 2005, pp. 1240--1245.

\bibitem{boria2005sensor}
F.~J. Boria, R.~J. Bachmann, P.~G. Ifju, R.~D. Quinn, R.~Vaidyanathan,
  C.~Perry, and J.~Wagener, ``A sensor platform capable of aerial and
  terrestrial locomotion,'' in \emph{2005 IEEE/RSJ International Conference on
  Intelligent Robots and Systems}.\hskip 1em plus 0.5em minus 0.4em\relax IEEE,
  2005, pp. 3959--3964.

\bibitem{peterson2011wing}
K.~Peterson, P.~Birkmeyer, R.~Dudley, and R.~Fearing, ``A wing-assisted running
  robot and implications for avian flight evolution,'' \emph{Bioinspiration \&
  biomimetics}, vol.~6, no.~4, p. 046008, 2011.

\bibitem{bachmann2009biologically}
R.~J. Bachmann, F.~J. Boria, R.~Vaidyanathan, P.~G. Ifju, and R.~D. Quinn, ``A
  biologically inspired micro-vehicle capable of aerial and terrestrial
  locomotion,'' \emph{Mechanism and Machine Theory}, vol.~44, no.~3, pp.
  513--526, 2009.

\bibitem{bachmann2009drive}
R.~J. Bachmann, R.~Vaidyanathan, and R.~D. Quinn, ``Drive train design enabling
  locomotion transition of a small hybrid air-land vehicle,'' in \emph{2009
  IEEE/RSJ International Conference on Intelligent Robots and Systems}.\hskip
  1em plus 0.5em minus 0.4em\relax IEEE, 2009, pp. 5647--5652.

\bibitem{yangjun2020progress}
Z.~Yangjun, Q.~Yuping, Z.~Weilin, Z.~Lei, P.~Jie, X.~Bin, and W.~Zexing,
  ``Progress and key technologies of flying cars,'' \emph{Journal of Automotive
  Safety and Energy}, vol.~11, no.~1, p.~1, 2020.

\bibitem{Shish2021}
K.~H. Shish, N.~Cramer, G.~Gorospe, T.~Lombaerts, K.~Kannan, and V.~Stepanyan,
  ``{Survey of capabilities and gaps in external perception sensors for
  autonomous urban air mobility applications},'' \emph{AIAA Scitech 2021
  Forum}, pp. 1--29, 2021.

\bibitem{Lombaerts2022}
T.~Lombaerts, K.~H. Shish, G.~Keller, V.~Stepanyan, N.~Cramer, and C.~Ippolito,
  ``{Adaptive Multi-Sensor Fusion Based Object Tracking for Autonomous Urban
  Air Mobility Operations},'' \emph{AIAA Science and Technology Forum and
  Exposition, AIAA SciTech Forum 2022}, 2022.

\bibitem{Garrow2021}
L.~A. Garrow, B.~J. German, and C.~E. Leonard, ``{Urban air mobility: A
  comprehensive review and comparative analysis with autonomous and electric
  ground transportation for informing future research},'' \emph{Transportation
  Research Part C: Emerging Technologies}, vol. 132, 2021.

\bibitem{nasaAdvancedMobility}
\BIBentryALTinterwordspacing
``{A}dvanced {A}ir {M}obility ({A}{A}{M}),'' {A}ccessed: 08-Jul-2022. [Online].
  Available: \url{https://www.nasa.gov/aam}
\BIBentrySTDinterwordspacing

\bibitem{ChinasMinistryofScienceandTechnology2022}
\BIBentryALTinterwordspacing
``{Medium- and Long-term Development Plan for Scientific and Technological
  Innovation in Transportation}". Accessed: 2022-07-08. [Online]. Available:
  \url{https://xxgk.mot.gov.cn/2020/jigou/kjs/202203/t20220325_3647752.html}
\BIBentrySTDinterwordspacing

\bibitem{Pan2021}
G.~Pan and M.~S. Alouini, ``{Flying Car Transportation System: Advances,
  Techniques, and Challenges},'' \emph{IEEE Access}, vol.~9, pp.
  24\,586--24\,603, 2021.

\bibitem{VINTAGEEVERYDAY2018}
\BIBentryALTinterwordspacing
``{Vintage Photos of 12 Cool Flying Cars That Really Existed in the Past}".
  Accessed: 2022-07-08. [Online]. Available:
  \url{https://www.vintag.es/2018/08/flying-cars.html}
\BIBentrySTDinterwordspacing

\bibitem{karnouskos2018self}
S.~Karnouskos, ``Self-driving car acceptance and the role of ethics,''
  \emph{IEEE Transactions on Engineering Management}, vol.~67, no.~2, pp.
  252--265, 2018.

\bibitem{Kleinman2021}
\BIBentryALTinterwordspacing
``{Flying Car Completes Test Flight between Airports}". Accessed: 2022-07-08.
  [Online]. Available: \url{https://www.bbc.co.uk/news/technology-57651843}
\BIBentrySTDinterwordspacing

\bibitem{PeterSigal2014}
\BIBentryALTinterwordspacing
``{Aeromobil 2.5}". Accessed: 2022-07-08. [Online]. Available:
  \url{https://www.nytimes.com/2014/08/24/automobiles/aeromobil-2-5.html}
\BIBentrySTDinterwordspacing

\bibitem{airbusItaldesignAirbus}
\BIBentryALTinterwordspacing
``{I}taldesign and {A}irbus unveil {P}op.{U}p". Accessed: 2022-07-08. [Online].
  Available:
  \url{https://www.airbus.com/en/newsroom/press-releases/2017-03-italdesign-and-airbus-unveil-popup}
\BIBentrySTDinterwordspacing

\bibitem{Salman2020}
M.~Salman, A.~Sameh, M.~Fanni, S.~Sugano, and A.~M. Mohamed, ``{Design,
  control, and dynamic simulation of securing and transformation mechanisms for
  a hybrid ground aerial robot},'' \emph{International Journal of Mechanical
  and Mechatronics Engineering}, vol.~20, no.~2, pp. 100--107, 2020.

\bibitem{fadhil2018gis}
D.~N. Fadhil, ``A gis-based analysis for selecting ground infrastructure
  locations for urban air mobility,'' \emph{inlangen]. Master’s Thesis,
  Technical University of Munich}, 2018.

\bibitem{pardede2020take}
W.~M. Pardede and M.~Adhitya, ``Take off and landing performance analysis for a
  flying car model using wind tunnel test method,'' in \emph{AIP Conference
  Proceedings}, vol. 2227, no.~1.\hskip 1em plus 0.5em minus 0.4em\relax AIP
  Publishing LLC, 2020, p. 020030.

\bibitem{Postorino2020}
M.~N. Postorino and G.~M. Sarn{\'{e}}, ``{Reinventing mobility paradigms:
  Flying car scenarios and challenges for urban mobility},''
  \emph{Sustainability (Switzerland)}, vol.~12, no.~9, pp. 1--16, 2020.

\bibitem{metinelectrically}
U.~Metin and S.~{\c{C}}oban, ``Electrically driven vtol flying car designing
  and aerodynamic analysis,'' \emph{Avrupa Bilim ve Teknoloji Dergisi}, no.~25,
  pp. 815--821.

\bibitem{rajashekara2016flying}
K.~Rajashekara, Q.~Wang, and K.~Matsuse, ``Flying cars: Challenges and
  propulsion strategies,'' \emph{IEEE Electrification Magazine}, vol.~4, no.~1,
  pp. 46--57, 2016.

\bibitem{BAESystems}
\BIBentryALTinterwordspacing
{BAE Systems}. ``{What is Avionics?}". Accessed: 2022-07-08. [Online].
  Available: \url{https://www.baesystems.com/en-us/definition/what-is-avionics}
\BIBentrySTDinterwordspacing

\bibitem{Cohen2021}
A.~P. Cohen, S.~A. Shaheen, and E.~M. Farrar, ``{Urban Air Mobility: History,
  Ecosystem, Market Potential, and Challenges},'' \emph{IEEE Transactions on
  Intelligent Transportation Systems}, vol.~22, no.~9, pp. 6074--6087, 2021.

\bibitem{goyal2018urban}
R.~Goyal, C.~Reiche, C.~Fernando, J.~Jacquie~Serrao, S.~Kimmel, A.~Cohen, and
  S.~Shaheen, ``Urban air mobility (uam) market study-booz allen hamilton
  technical briefing,'' Tech. Rep., 2018.

\bibitem{goyal2021advanced}
R.~Goyal, C.~Reiche, C.~Fernando, and A.~Cohen, ``Advanced air mobility: Demand
  analysis and market potential of the airport shuttle and air taxi markets,''
  \emph{Sustainability}, vol.~13, no.~13, p. 7421, 2021.

\bibitem{grandl2018future}
G.~Grandl, M.~Ostgathe, J.~Cachay, S.~Doppler, J.~Salib, H.~Ross, J.~Detert,
  and R.~Kallenberg, ``The future of vertical mobility sizing the market for
  passenger, inspection, and goods services until 2035,'' \emph{Porsche
  Consulting. https://fedotov.
  co/wp-content/uploads/2018/03/Future-of-Vertical-Mobility. pdf}, 2018.

\bibitem{wang2014autobody}
Y.~F. Wang and T.~X. Su, ``Autobody modelling analysis of the flying cars,'' in
  \emph{Applied Mechanics and Materials}, vol. 577.\hskip 1em plus 0.5em minus
  0.4em\relax Trans Tech Publ, 2014, pp. 1310--1313.

\bibitem{Straubinger2020}
A.~Straubinger, R.~Rothfeld, M.~Shamiyeh, K.~D. B{\"{u}}chter, J.~Kaiser, and
  K.~O. Pl{\"{o}}tner, ``{An overview of current research and developments in
  urban air mobility – Setting the scene for UAM introduction},''
  \emph{Journal of Air Transport Management}, vol.~87, 2020.

\bibitem{BBC}
\BIBentryALTinterwordspacing
``{Dubai Announces Passenger Drone Plans - BBC News}". Accessed: 2022-07-08.
  [Online]. Available: \url{https://www.bbc.com/news/technology-38967235}
\BIBentrySTDinterwordspacing

\bibitem{2020An}
A.~Straubinger, R.~Rothfeld, M.~Shamiyeh, K.~Büchter, and K.~O. Pltner, ``An
  overview of current research and developments in urban air mobility –
  setting the scene for uam introduction,'' \emph{Journal of Air Transport
  Management}, vol.~87, p. 101852, 2020.

\bibitem{luo2021simulation}
Y.~Luo, Y.~Qian, Z.~Zeng, and Y.~Zhang, ``Simulation and analysis of operating
  characteristics of power battery for flying car utilization,''
  \emph{eTransportation}, vol.~8, p. 100111, 2021.

\bibitem{silva2018vtol}
C.~Silva, W.~R. Johnson, E.~Solis, M.~D. Patterson, and K.~R. Antcliff, ``Vtol
  urban air mobility concept vehicles for technology development,'' in
  \emph{2018 Aviation Technology, Integration, and Operations Conference},
  2018, p. 3847.

\bibitem{Vempati2021}
L.~Vempati, M.~Geffard, and A.~Anderegg, ``{Assessing Human-Automation Role
  Challenges for Urban Air Mobility (UAM) Operations},'' \emph{AIAA/IEEE
  Digital Avionics Systems Conference - Proceedings}, vol. 2021-October, 2021.

\bibitem{Kelley2022}
B.~N. Kelley, W.~J. Waltz, A.~Miloslavsky, R.~Williams, A.~K. Ishihara,
  I.~Reyes, and H.~R. Idris, ``{Designing A Distributed Web-based Simulation
  Environment for Enabling Autonomous Systems Research},'' \emph{AIAA Science
  and Technology Forum and Exposition, AIAA SciTech Forum 2022}, 2022.

\bibitem{erturk2020requirements}
M.~C. Ert{\"u}rk, N.~Hosseini, H.~Jamal, A.~{\c{S}}ahin, D.~Matolak, and
  J.~Haque, ``Requirements and technologies towards uam: Communication,
  navigation, and surveillance,'' in \emph{2020 Integrated Communications
  Navigation and Surveillance Conference (ICNS)}.\hskip 1em plus 0.5em minus
  0.4em\relax IEEE, 2020, pp. 2C2--1.

\bibitem{zeng2021performance}
T.~Zeng, O.~Semiari, W.~Saad, and M.~Bennis, ``Performance analysis of
  aircraft-to-ground communication networks in urban air mobility (uam),'' in
  \emph{2021 IEEE Global Communications Conference (GLOBECOM)}.\hskip 1em plus
  0.5em minus 0.4em\relax IEEE, 2021, pp. 1--6.

\bibitem{kenney2011dedicated}
J.~B. Kenney, ``Dedicated short-range communications (dsrc) standards in the
  united states,'' \emph{Proceedings of the IEEE}, vol.~99, no.~7, pp.
  1162--1182, 2011.

\bibitem{ahmad2019lte}
I.~Ahmad, R.~Md~Noor, and M.~Reza~Z'aba, ``Lte efficiency when used in traffic
  information systems: A stable interest aware clustering,''
  \emph{International Journal of Communication Systems}, vol.~32, no.~2, p.
  e3853, 2019.

\bibitem{federal2010second}
F.~C. Commission \emph{et~al.}, ``Second memorandum opinion and order,''
  \emph{FCC 10-174}, 2010.

\bibitem{azari2019cellular}
M.~M. Azari, F.~Rosas, and S.~Pollin, ``Cellular connectivity for uavs: Network
  modeling, performance analysis, and design guidelines,'' \emph{IEEE
  Transactions on Wireless Communications}, vol.~18, no.~7, pp. 3366--3381,
  2019.

\bibitem{saeed2021wireless}
N.~Saeed, T.~Y. Al-Naffouri, and M.-S. Alouini, ``Wireless communication for
  flying cars,'' \emph{Frontiers in Communications and Networks}, vol.~2,
  p.~16, 2021.

\bibitem{alsamhi2019tethered}
S.~Alsamhi, M.~S. Ansari, L.~Zhao, S.~N. Van, S.~Gupta, A.~A. Alammari, A.~H.
  Saber, M.~Y. Hebah, M.~A.~A. Alasali, H.~M. Aljabali \emph{et~al.},
  ``Tethered balloon technology for green communication in smart cities and
  healthy environment,'' in \emph{2019 First International Conference of
  Intelligent Computing and Engineering (ICOICE)}.\hskip 1em plus 0.5em minus
  0.4em\relax IEEE, 2019, pp. 1--7.

\bibitem{tozer2001high}
T.~Tozer and D.~Grace, ``High-altitude platforms for wireless communications,''
  \emph{Electronics \& communication engineering journal}, vol.~13, no.~3, pp.
  127--137, 2001.

\bibitem{lu2018survey}
Y.~Lu, Z.~Xue, G.-S. Xia, and L.~Zhang, ``A survey on vision-based uav
  navigation,'' \emph{Geo-spatial information science}, vol.~21, no.~1, pp.
  21--32, 2018.

\bibitem{araar2014new}
O.~Araar and N.~Aouf, ``A new hybrid approach for the visual servoing of vtol
  uavs from unknown geometries,'' in \emph{22nd Mediterranean Conference on
  Control and Automation}.\hskip 1em plus 0.5em minus 0.4em\relax IEEE, 2014,
  pp. 1425--1432.

\bibitem{cesetti2010vision}
A.~Cesetti, E.~Frontoni, A.~Mancini, P.~Zingaretti, and S.~Longhi, ``A
  vision-based guidance system for uav navigation and safe landing using
  natural landmarks,'' \emph{Journal of intelligent and robotic systems},
  vol.~57, no.~1, pp. 233--257, 2010.

\bibitem{carrillo2012combining}
L.~R.~G. Carrillo, A.~E.~D. L{\'o}pez, R.~Lozano, and C.~P{\'e}gard,
  ``Combining stereo vision and inertial navigation system for a quad-rotor
  uav,'' \emph{Journal of intelligent \& robotic systems}, vol.~65, no.~1, pp.
  373--387, 2012.

\bibitem{ho2018adaptive}
H.~Ho, G.~C. de~Croon, E.~Van~Kampen, Q.~Chu, and M.~Mulder, ``Adaptive gain
  control strategy for constant optical flow divergence landing,'' \emph{IEEE
  Transactions on Robotics}, vol.~34, no.~2, pp. 508--516, 2018.

\bibitem{minaee2021image}
S.~Minaee, Y.~Y. Boykov, F.~Porikli, A.~J. Plaza, N.~Kehtarnavaz, and
  D.~Terzopoulos, ``Image segmentation using deep learning: A survey,''
  \emph{IEEE Transactions on Pattern Analysis and Machine Intelligence}, 2021.

\bibitem{osco2021review}
L.~P. Osco, J.~M. Junior, A.~P.~M. Ramos, L.~A. de~Castro~Jorge, S.~N.
  Fatholahi, J.~de~Andrade~Silva, E.~T. Matsubara, H.~Pistori, W.~N.
  Gon{\c{c}}alves, and J.~Li, ``A review on deep learning in uav remote
  sensing,'' \emph{International Journal of Applied Earth Observation and
  Geoinformation}, vol. 102, p. 102456, 2021.

\bibitem{kirillov2019panoptic}
A.~Kirillov, K.~He, R.~Girshick, C.~Rother, and P.~Doll{\'a}r, ``Panoptic
  segmentation,'' in \emph{Proceedings of the IEEE/CVF Conference on Computer
  Vision and Pattern Recognition}, 2019, pp. 9404--9413.

\bibitem{Lyu2021}
\BIBentryALTinterwordspacing
Y.~Lyu, G.~Vosselman, G.-S. Xia, and M.~Y. Yang, ``{Bidirectional Multi-scale
  Attention Networks for Semantic Segmentation of Oblique UAV Imagery},'' 2021.
  [Online]. Available: \url{http://arxiv.org/abs/2102.03099}
\BIBentrySTDinterwordspacing

\bibitem{Wallace2012}
L.~Wallace, A.~Lucieer, C.~Watson, and D.~Turner, ``{Development of a UAV-LiDAR
  system with application to forest inventory},'' \emph{Remote Sensing},
  vol.~4, no.~6, pp. 1519--1543, 2012.

\bibitem{Khan2017}
S.~Khan, L.~Arag{\~{a}}o, and J.~Iriarte, ``{A UAV–lidar system to map
  Amazonian rainforest and its ancient landscape transformations},''
  \emph{International Journal of Remote Sensing}, vol.~38, no. 8-10, pp.
  2313--2330, 2017.

\bibitem{Fuad2018}
N.~A. Fuad, Z.~Ismail, Z.~Majid, N.~Darwin, M.~F. Ariff, K.~M. Idris, and A.~R.
  Yusoff, ``{Accuracy evaluation of digital terrain model based on different
  flying altitudes and conditional of terrain using UAV LiDAR technology},''
  \emph{IOP Conference Series: Earth and Environmental Science}, vol. 169,
  no.~1, 2018.

\bibitem{Lin2019}
Y.~C. Lin, Y.~T. Cheng, T.~Zhou, R.~Ravi, S.~M. Hasheminasab, J.~E. Flatt,
  C.~Troy, and A.~Habib, ``{Evaluation of UAV LiDAR for mapping coastal
  environments},'' \emph{Remote Sensing}, vol.~11, no.~24, 2019.

\bibitem{yan2018automated}
W.~Yan, H.~Guan, L.~Cao, Y.~Yu, S.~Gao, and J.~Lu, ``An automated hierarchical
  approach for three-dimensional segmentation of single trees using uav lidar
  data,'' \emph{Remote Sensing}, vol.~10, no.~12, p. 1999, 2018.

\bibitem{Hayton2020}
J.~N. Hayton, T.~Barros, C.~Premebida, M.~J. Coombes, and U.~J. Nunes,
  ``{CNN-based Human Detection Using a 3D LiDAR onboard a UAV},'' \emph{2020
  IEEE International Conference on Autonomous Robot Systems and Competitions,
  ICARSC 2020}, pp. 312--318, 2020.

\bibitem{Scherer2012}
S.~Scherer, L.~Chamberlain, and S.~Singh, ``{Autonomous landing at unprepared
  sites by a full-scale helicopter},'' \emph{Robotics and Autonomous Systems},
  vol.~60, no.~12, pp. 1545--1562, 2012.

\bibitem{Cazzato2020}
D.~Cazzato, C.~Cimarelli, J.~L. Sanchez-Lopez, H.~Voos, and M.~Leo, ``{A survey
  of computer vision methods for 2d object detection from unmanned aerial
  vehicles},'' \emph{Journal of Imaging}, vol.~6, no.~8, 2020.

\bibitem{Jin2019}
R.~Jin, H.~M. Owais, D.~Lin, T.~Song, and Y.~Yuan, ``{Ellipse proposal and
  convolutional neural network discriminant for autonomous landing marker
  detection},'' \emph{Journal of Field Robotics}, vol.~36, no.~1, pp. 6--16,
  2019.

\bibitem{de2015board}
F.~De~Smedt, D.~Hulens, and T.~Goedem{\'e}, ``On-board real-time tracking of
  pedestrians on a uav,'' in \emph{Proceedings of the IEEE conference on
  computer vision and pattern recognition workshops}, 2015, pp. 1--8.

\bibitem{DeSmedt2015}
F.~{De Smedt}, D.~Hulens, and T.~Goedeme, ``{On-board real-time tracking of
  pedestrians on a UAV},'' \emph{IEEE Computer Society Conference on Computer
  Vision and Pattern Recognition Workshops}, vol. 2015-Octob, pp. 1--8, 2015.

\bibitem{fu2020dr}
C.~Fu, F.~Ding, Y.~Li, J.~Jin, and C.~Feng, ``Dr 2 track: Towards real-time
  visual tracking for uav via distractor repressed dynamic regression,'' in
  \emph{2020 IEEE/RSJ International Conference on Intelligent Robots and
  Systems (IROS)}.\hskip 1em plus 0.5em minus 0.4em\relax IEEE, 2020, pp.
  1597--1604.

\bibitem{li2022all}
B.~Li, C.~Fu, F.~Ding, J.~Ye, and F.~Lin, ``All-day object tracking for
  unmanned aerial vehicle,'' \emph{IEEE Transactions on Mobile Computing},
  2022.

\bibitem{Mueller2016}
M.~Mueller, N.~Smith, and B.~Ghanem, ``{A benchmark and simulator for UAV
  tracking},'' \emph{Lecture Notes in Computer Science (including subseries
  Lecture Notes in Artificial Intelligence and Lecture Notes in
  Bioinformatics)}, vol. 9905 LNCS, pp. 445--461, 2016.

\bibitem{Li2017}
S.~Li and D.~Y. Yeung, ``{Visual object tracking for unmanned aerial vehicles:
  A benchmark and new motion models},'' \emph{31st AAAI Conference on
  Artificial Intelligence, AAAI 2017}, pp. 4140--4146, 2017.

\bibitem{yu2020unmanned}
H.~Yu, G.~Li, W.~Zhang, Q.~Huang, D.~Du, Q.~Tian, and N.~Sebe, ``The unmanned
  aerial vehicle benchmark: Object detection, tracking and baseline,''
  \emph{International Journal of Computer Vision}, vol. 128, no.~5, pp.
  1141--1159, 2020.

\bibitem{Nigam2018}
I.~Nigam, C.~Huang, and D.~Ramanan, ``{Ensemble Knowledge Transfer for Semantic
  Segmentation},'' \emph{Proceedings - 2018 IEEE Winter Conference on
  Applications of Computer Vision, WACV 2018}, vol. 2018-Janua, pp. 1499--1508,
  2018.

\bibitem{zhu2018vision}
P.~Zhu, L.~Wen, X.~Bian, H.~Ling, and Q.~Hu, ``Vision meets drones: A
  challenge,'' \emph{arXiv preprint arXiv:1804.07437}, 2018.

\bibitem{Fonder2019}
M.~Fonder and M.~{Van Droogenbroeck}, ``{Mid-air: A multi-modal dataset for
  extremely low altitude drone flights},'' \emph{IEEE Computer Society
  Conference on Computer Vision and Pattern Recognition Workshops}, vol.
  2019-June, pp. 553--562, 2019.

\bibitem{Lyu2020}
Y.~Lyu, G.~Vosselman, G.~S. Xia, A.~Yilmaz, and M.~Y. Yang, ``{UAVid: A
  semantic segmentation dataset for UAV imagery},'' \emph{ISPRS Journal of
  Photogrammetry and Remote Sensing}, vol. 165, pp. 108--119, 2020.

\bibitem{bozcan2020air}
I.~Bozcan and E.~Kayacan, ``Au-air: A multi-modal unmanned aerial vehicle
  dataset for low altitude traffic surveillance,'' in \emph{2020 IEEE
  International Conference on Robotics and Automation (ICRA)}.\hskip 1em plus
  0.5em minus 0.4em\relax IEEE, 2020, pp. 8504--8510.

\bibitem{hu2021towards}
Q.~Hu, B.~Yang, S.~Khalid, W.~Xiao, N.~Trigoni, and A.~Markham, ``Towards
  semantic segmentation of urban-scale 3d point clouds: A dataset, benchmarks
  and challenges,'' in \emph{Proceedings of the IEEE/CVF Conference on Computer
  Vision and Pattern Recognition}, 2021, pp. 4977--4987.

\bibitem{li2020campus3d}
X.~Li, C.~Li, Z.~Tong, A.~Lim, J.~Yuan, Y.~Wu, J.~Tang, and R.~Huang,
  ``Campus3d: A photogrammetry point cloud benchmark for hierarchical
  understanding of outdoor scene,'' in \emph{Proceedings of the 28th ACM
  International Conference on Multimedia}, 2020, pp. 238--246.

\bibitem{tan2020toronto}
W.~Tan, N.~Qin, L.~Ma, Y.~Li, J.~Du, G.~Cai, K.~Yang, and J.~Li, ``Toronto-3d:
  A large-scale mobile lidar dataset for semantic segmentation of urban
  roadways,'' in \emph{Proceedings of the IEEE/CVF Conference on Computer
  Vision and Pattern Recognition Workshops}, 2020, pp. 202--203.

\bibitem{mueller2016benchmark}
M.~Mueller, N.~Smith, and B.~Ghanem, ``A benchmark and simulator for uav
  tracking,'' in \emph{European conference on computer vision}.\hskip 1em plus
  0.5em minus 0.4em\relax Springer, 2016, pp. 445--461.

\bibitem{bosch2019captioning}
M.~Bosch, C.~Gifford, A.~Ciesielski, S.~Almes, R.~Ellison, and G.~Christie,
  ``Captioning of full motion video from unmanned aerial platforms,'' in
  \emph{Geospatial Informatics IX}, vol. 10992.\hskip 1em plus 0.5em minus
  0.4em\relax International Society for Optics and Photonics, 2019, p. 1099202.

\bibitem{wang2014study}
Y.~F. Wang, T.~X. Su, and S.~W. Yang, ``Study on dynamic performance of a
  flying car body frame based on ansys,'' in \emph{Applied Mechanics and
  Materials}, vol. 602.\hskip 1em plus 0.5em minus 0.4em\relax Trans Tech Publ,
  2014, pp. 163--166.

\bibitem{zhao2018survey}
Y.~Zhao, Z.~Zheng, and Y.~Liu, ``Survey on computational-intelligence-based uav
  path planning,'' \emph{Knowledge-Based Systems}, vol. 158, pp. 54--64, 2018.

\bibitem{straubinger2020overview}
A.~Straubinger, R.~Rothfeld, M.~Shamiyeh, K.-D. B{\"u}chter, J.~Kaiser, and
  K.~O. Pl{\"o}tner, ``An overview of current research and developments in
  urban air mobility--setting the scene for uam introduction,'' \emph{Journal
  of Air Transport Management}, vol.~87, p. 101852, 2020.

\bibitem{sharif2019new}
A.~Sharif, S.~Choi, and H.~Roth, ``A new algorithm for autonomous outdoor
  navigation of robots that can fly and drive,'' in \emph{Proceedings of the
  5th International Conference on Mechatronics and Robotics Engineering}, 2019,
  pp. 141--145.

\bibitem{TerrySuh2020}
H.~J. {Terry Suh}, X.~Xiong, A.~Singletary, A.~D. Ames, and J.~W. Burdick,
  ``{Energy-efficient motion planning for multi-modal hybrid locomotion},''
  \emph{IEEE International Conference on Intelligent Robots and Systems}, pp.
  7027--7033, 2020.

\bibitem{sharif2018energy}
A.~Sharif, H.~Lahiru, S.~Herath, and H.~Roth, ``Energy efficient path planning
  of hybrid fly-drive robot (hyfdr) using a* algorithm.'' in \emph{ICINCO (2)},
  2018, pp. 211--220.

\bibitem{suh2020energy}
H.~T. Suh, X.~Xiong, A.~Singletary, A.~D. Ames, and J.~W. Burdick,
  ``Energy-efficient motion planning for multi-modal hybrid locomotion,'' in
  \emph{2020 IEEE/RSJ International Conference on Intelligent Robots and
  Systems (IROS)}.\hskip 1em plus 0.5em minus 0.4em\relax IEEE, 2020, pp.
  7027--7033.

\bibitem{choudhury2019dynamic}
S.~Choudhury, J.~P. Knickerbocker, and M.~J. Kochenderfer, ``Dynamic real-time
  multimodal routing with hierarchical hybrid planning,'' in \emph{2019 IEEE
  Intelligent Vehicles Symposium (IV)}.\hskip 1em plus 0.5em minus 0.4em\relax
  IEEE, 2019, pp. 2397--2404.

\bibitem{Zhang2013}
Y.~Zhang, L.~Wu, and S.~Wang, ``{UCAV path planning by Fitness-scaling Adaptive
  Chaotic Particle Swarm Optimization},'' \emph{Mathematical Problems in
  Engineering}, vol. 2013, 2013.

\bibitem{Liu2017}
J.~Liu, J.~Yang, H.~Liu, X.~Tian, and M.~Gao, ``{An improved ant colony
  algorithm for robot path planning},'' \emph{Soft Computing}, vol.~21, no.~19,
  pp. 5829--5839, 2017.

\bibitem{Sahingoz2013}
O.~K. Sahingoz, ``{Flyable path planning for a multi-UAV system with Genetic
  Algorithms and Bezier curves},'' \emph{2013 International Conference on
  Unmanned Aircraft Systems, ICUAS 2013 - Conference Proceedings}, pp. 41--48,
  2013.

\bibitem{Ran2011}
M.~Ran, H.~Duan, X.~Gao, and Z.~Mao, ``{Improved particle swarm optimization
  approach to path planning of amphibious mouse robot},'' \emph{Proceedings of
  the 2011 6th IEEE Conference on Industrial Electronics and Applications,
  ICIEA 2011}, pp. 1146--1149, 2011.

\bibitem{subosits2019racetrack}
J.~K. Subosits and J.~C. Gerdes, ``From the racetrack to the road: Real-time
  trajectory replanning for autonomous driving,'' \emph{IEEE Transactions on
  Intelligent Vehicles}, vol.~4, no.~2, pp. 309--320, 2019.

\bibitem{Loureiro2020}
G.~Loureiro, A.~Dias, and A.~Martins, ``{Survey of approaches for emergency
  landing spot detection with unmanned aerial vehicles},'' \emph{Robots in
  Human Life- Proceedings of the 23rd International Conference on Climbing and
  Walking Robots and the Support Technologies for Mobile Machines, CLAWAR
  2020}, pp. 129--136, 2020.

\bibitem{josef2020deep}
S.~Josef and A.~Degani, ``Deep reinforcement learning for safe local planning
  of a ground vehicle in unknown rough terrain,'' \emph{IEEE Robotics and
  Automation Letters}, vol.~5, no.~4, pp. 6748--6755, 2020.

\bibitem{yan2020towards}
C.~Yan, X.~Xiang, and C.~Wang, ``Towards real-time path planning through deep
  reinforcement learning for a uav in dynamic environments,'' \emph{Journal of
  Intelligent \& Robotic Systems}, vol.~98, no.~2, pp. 297--309, 2020.

\bibitem{theile2020uav}
M.~Theile, H.~Bayerlein, R.~Nai, D.~Gesbert, and M.~Caccamo, ``Uav coverage
  path planning under varying power constraints using deep reinforcement
  learning,'' in \emph{2020 IEEE/RSJ International Conference on Intelligent
  Robots and Systems (IROS)}.\hskip 1em plus 0.5em minus 0.4em\relax IEEE,
  2020, pp. 1444--1449.

\bibitem{hubmann2018automated}
C.~Hubmann, J.~Schulz, M.~Becker, D.~Althoff, and C.~Stiller, ``Automated
  driving in uncertain environments: Planning with interaction and uncertain
  maneuver prediction,'' \emph{IEEE transactions on intelligent vehicles},
  vol.~3, no.~1, pp. 5--17, 2018.

\bibitem{di2017evaluating}
P.~F. Di~Donato and E.~M. Atkins, ``Evaluating risk to people and property for
  aircraft emergency landing planning,'' \emph{Journal of Aerospace Information
  Systems}, vol.~14, no.~5, pp. 259--278, 2017.

\bibitem{ayhan2018semi}
B.~Ayhan, C.~Kwan, Y.-B. Um, B.~Budavari, and J.~Larkin, ``Semi-automated
  emergency landing site selection approach for uavs,'' \emph{IEEE Transactions
  on Aerospace and Electronic Systems}, vol.~55, no.~4, pp. 1892--1906, 2018.

\bibitem{Coombes2016}
M.~Coombes, W.~H. Chen, and P.~Render, ``{Site selection during unmanned aerial
  system forced landings using decision-making Bayesian networks},''
  \emph{Journal of Aerospace Information Systems}, vol.~13, no.~12, pp.
  491--495, 2016.

\bibitem{loureiro2020survey}
G.~Loureiro, A.~Dias, and A.~Martins, ``Survey of approaches for emergency
  landing spot detection with unmanned aerial vehicles,'' \emph{Proceedings of
  the Robots in Human Life—CLAWAR}, pp. 129--136, 2020.

\bibitem{lusk2018vision}
P.~C. Lusk, ``Vision-based emergency landing of small unmanned aircraft
  systems,'' Ph.D. dissertation, Brigham Young University, 2018.

\bibitem{Arantes2018}
J.~D.~S. Arantes, M.~D.~S. Arantes, A.~B. Missaglia, E.~D.~V. Simoes, and C.~F.
  {Motta Toledo}, ``{Evaluating hardware platforms and path re-planning
  strategies for the uav emergency landing problem},'' \emph{Proceedings -
  International Conference on Tools with Artificial Intelligence, ICTAI}, vol.
  2017-Novem, pp. 937--944, 2018.

\bibitem{gonzalez2021visual}
J.~Gonz{\'a}lez-Trejo, D.~Mercado-Ravell, I.~Becerra, and R.~Murrieta-Cid, ``On
  the visual-based safe landing of uavs in populated areas: a crucial aspect
  for urban deployment,'' \emph{IEEE Robotics and Automation Letters}, vol.~6,
  no.~4, pp. 7901--7908, 2021.

\bibitem{Mintchev2018}
S.~Mintchev and D.~Floreano, ``{A multi-modal hovering and terrestrial robot
  with adaptive morphology},'' \emph{2nd International Symposium on Aerial
  Robotics}, 2018.

\bibitem{Meiri2019}
N.~Meiri and D.~Zarrouk, ``{Flying STAR, a hybrid crawling and flying sprawl
  tuned robot},'' \emph{Proceedings - IEEE International Conference on Robotics
  and Automation}, vol. 2019-May, pp. 5302--5308, 2019.

\bibitem{kim2013study}
K.~Kim, K.~Hwang, and H.~Kim, ``Study of an adaptive fuzzy algorithm to control
  a rectangular-shaped unmanned surveillance flying car,'' \emph{Journal of
  Mechanical Science and Technology}, vol.~27, no.~8, pp. 2477--2486, 2013.

\bibitem{paden2016survey}
B.~Paden, M.~{\v{C}}{\'a}p, S.~Z. Yong, D.~Yershov, and E.~Frazzoli, ``A survey
  of motion planning and control techniques for self-driving urban vehicles,''
  \emph{IEEE Transactions on intelligent vehicles}, vol.~1, no.~1, pp. 33--55,
  2016.

\bibitem{jiang2018lateral}
J.~Jiang and A.~Astolfi, ``Lateral control of an autonomous vehicle,''
  \emph{IEEE Transactions on Intelligent Vehicles}, vol.~3, no.~2, pp.
  228--237, 2018.

\bibitem{tan2021multimodal}
Q.~Tan, X.~Zhang, H.~Liu, S.~Jiao, M.~Zhou, and J.~Li, ``Multimodal dynamics
  analysis and control for amphibious fly-drive vehicle,'' \emph{IEEE/ASME
  Transactions on Mechatronics}, vol.~26, no.~2, pp. 621--632, 2021.

\bibitem{keshavarzian2020pso}
H.~Keshavarzian and K.~Daneshjou, ``Pso-based online estimation of aerodynamic
  ground effect in the backstepping controller of quadrotor,'' \emph{Journal of
  the Brazilian Society of Mechanical Sciences and Engineering}, vol.~42,
  no.~11, pp. 1--10, 2020.

\bibitem{matus2021ground}
A.~Matus-Vargas, G.~Rodriguez-Gomez, and J.~Martinez-Carranza, ``Ground effect
  on rotorcraft unmanned aerial vehicles: a review,'' \emph{Intelligent Service
  Robotics}, pp. 1--20, 2021.

\bibitem{griffiths2002study}
D.~A. Griffiths, ``A study of dual-rotor interference and ground effect using a
  free-vortex wake model,'' in \emph{American Helicopter Society 58th Annual
  Forum, Montreal, Canada, June 11-13, 2002}, 2002.

\bibitem{curtiss1984rotor}
H.~Curtiss, M.~Sun, W.~Putman, and E.~Hanker, ``Rotor aerodynamics in ground
  effect at low advance ratios,'' \emph{Journal of the American Helicopter
  Society}, vol.~29, no.~1, pp. 48--55, 1984.

\bibitem{krishnan2016numerical}
P.~Krishnan~Rajendran, ``Numerical investigation of aerodynamic characteristics
  of flying car,'' Ph.D. dissertation, Instytut Techniki Lotniczej i Mechaniki
  Stosowanej, 2016.

\bibitem{cheeseman1955effect}
I.~Cheeseman and W.~Bennett, ``The effect of the ground on a helicopter rotor
  in forward flight,'' 1955.

\bibitem{Bernard2017}
D.~D.~C. Bernard, F.~Riccardi, M.~Giurato, and M.~Lovera, ``{A dynamic analysis
  of ground effect for a quadrotor platform},'' \emph{IFAC-PapersOnLine},
  vol.~50, no.~1, pp. 10\,311--10\,316, 2017.

\bibitem{Conyers2018}
S.~A. Conyers, M.~J. Rutherford, and K.~P. Valavanis, ``{An empirical
  evaluation of ground effect for small-scale rotorcraft},'' \emph{Proceedings
  - IEEE International Conference on Robotics and Automation}, pp. 1244--1250,
  2018.

\bibitem{Keshavarzian2020}
H.~Keshavarzian and K.~Daneshjou, ``{PSO-based online estimation of aerodynamic
  ground effect in the backstepping controller of quadrotor},'' \emph{Journal
  of the Brazilian Society of Mechanical Sciences and Engineering}, vol.~42,
  no.~11, 2020.

\bibitem{Keshavarzian2019}
------, ``{Modified under-actuated quadrotor model for forwarding flight in the
  presence of ground effect},'' \emph{Aerospace Science and Technology},
  vol.~89, pp. 242--252, 2019.

\bibitem{Sanchez-Cuevas2017}
P.~Sanchez-Cuevas, G.~Heredia, and A.~Ollero, ``{Characterization of the
  aerodynamic ground effect and its influence in multirotor control},''
  \emph{International Journal of Aerospace Engineering}, vol. 2017, 2017.

\bibitem{yang2022neuroadaptive}
G.~Yang, J.~Yao, and N.~Ullah, ``Neuroadaptive control of saturated nonlinear
  systems with disturbance compensation,'' \emph{ISA transactions}, vol. 122,
  pp. 49--62, 2022.

\bibitem{yang2022neuroadaptive1}
G.~Yang, J.~Yao, and Z.~Dong, ``Neuroadaptive learning algorithm for
  constrained nonlinear systems with disturbance rejection,''
  \emph{International Journal of Robust and Nonlinear Control}, 2022.

\bibitem{courtin2018feasibility}
C.~Courtin, M.~J. Burton, A.~Yu, P.~Butler, P.~D. Vascik, and R.~J. Hansman,
  ``Feasibility study of short takeoff and landing urban air mobility vehicles
  using geometric programming,'' in \emph{2018 Aviation Technology,
  Integration, and Operations Conference}, 2018, p. 4151.

\bibitem{openerOpeneraero}
\BIBentryALTinterwordspacing
Opener. Accessed: 2022-07-08. [Online]. Available: \url{https://opener.aero/}
\BIBentrySTDinterwordspacing

\bibitem{teslaTransitioningTesla}
\BIBentryALTinterwordspacing
Tesla. ``{T}ransitioning to {T}esla {V}ision". Accessed: 2022-07-08. [Online].
  Available: \url{https://www.tesla.com/support/transitioning-tesla-vision}
\BIBentrySTDinterwordspacing

\bibitem{newatlasAutonomousRace}
\BIBentryALTinterwordspacing
``{A}utonomous race car sets new self-driving land speed record. Accessed
  2022-07-08. [Online]. Available:
  \url{https://newatlas.com/automotive/autonomous-land-speed-record/}
\BIBentrySTDinterwordspacing

\bibitem{kabzan2020amz}
J.~Kabzan, M.~I. Valls, V.~J. Reijgwart, H.~F. Hendrikx, C.~Ehmke, M.~Prajapat,
  A.~B{\"u}hler, N.~Gosala, M.~Gupta, R.~Sivanesan \emph{et~al.}, ``Amz
  driverless: The full autonomous racing system,'' \emph{Journal of Field
  Robotics}, vol.~37, no.~7, pp. 1267--1294, 2020.

\bibitem{kabzan2019learning}
J.~Kabzan, L.~Hewing, A.~Liniger, and M.~N. Zeilinger, ``Learning-based model
  predictive control for autonomous racing,'' \emph{IEEE Robotics and
  Automation Letters}, vol.~4, no.~4, pp. 3363--3370, 2019.

\bibitem{weiss2020deepracing}
T.~Weiss and M.~Behl, ``Deepracing: a framework for autonomous racing,'' in
  \emph{2020 Design, Automation \& Test in Europe Conference \& Exhibition
  (DATE)}.\hskip 1em plus 0.5em minus 0.4em\relax IEEE, 2020, pp. 1163--1168.

\bibitem{sada2018image}
M.~M. Sada and M.~Mahesh, ``Image deblurring techniques—a detail review,''
  \emph{Int. J. Sci. Res. Sci. Eng. Technol}, vol.~4, no.~2, p.~15, 2018.

\bibitem{nah2021ntire}
S.~Nah, S.~Son, S.~Lee, R.~Timofte, and K.~M. Lee, ``Ntire 2021 challenge on
  image deblurring,'' in \emph{Proceedings of the IEEE/CVF Conference on
  Computer Vision and Pattern Recognition}, 2021, pp. 149--165.

\bibitem{schuler2016learning}
C.~J. Schuler, M.~Hirsch, S.~Harmeling, and B.~Scholkopf, ``Learning to
  deblur,'' \emph{IEEE Transactions on Pattern Analysis \& Machine
  Intelligence}, vol.~38, no.~07, pp. 1439--1451, 2016.

\bibitem{hu2014joint}
Z.~Hu, L.~Xu, and M.-H. Yang, ``Joint depth estimation and camera shake removal
  from single blurry image,'' in \emph{Proceedings of the IEEE Conference on
  Computer Vision and Pattern Recognition}, 2014, pp. 2893--2900.

\bibitem{paramanand2013non}
C.~Paramanand and A.~N. Rajagopalan, ``Non-uniform motion deblurring for
  bilayer scenes,'' in \emph{Proceedings of the IEEE Conference on Computer
  Vision and Pattern Recognition}, 2013, pp. 1115--1122.

\bibitem{hyun2013dynamic}
T.~Hyun~Kim, B.~Ahn, and K.~Mu~Lee, ``Dynamic scene deblurring,'' in
  \emph{Proceedings of the IEEE international conference on computer vision},
  2013, pp. 3160--3167.

\bibitem{nah2017deep}
S.~Nah, T.~Hyun~Kim, and K.~Mu~Lee, ``Deep multi-scale convolutional neural
  network for dynamic scene deblurring,'' in \emph{Proceedings of the IEEE
  conference on computer vision and pattern recognition}, 2017, pp. 3883--3891.

\bibitem{noroozi2017motion}
M.~Noroozi, P.~Chandramouli, and P.~Favaro, ``Motion deblurring in the wild,''
  in \emph{German conference on pattern recognition}.\hskip 1em plus 0.5em
  minus 0.4em\relax Springer, 2017, pp. 65--77.

\bibitem{zhang2018gated}
X.~Zhang, H.~Dong, Z.~Hu, W.-S. Lai, F.~Wang, and M.-H. Yang, ``Gated fusion
  network for joint image deblurring and super-resolution,'' \emph{arXiv
  preprint arXiv:1807.10806}, 2018.

\bibitem{goodfellow2014generative}
I.~Goodfellow, J.~Pouget-Abadie, M.~Mirza, B.~Xu, D.~Warde-Farley, S.~Ozair,
  A.~Courville, and Y.~Bengio, ``Generative adversarial nets,'' \emph{Advances
  in neural information processing systems}, vol.~27, 2014.

\bibitem{wei2021dynamic}
B.~Wei, L.~Zhang, K.~Wang, Q.~Kong, and Z.~Wang, ``Dynamic scene deblurring and
  image de-raining based on generative adversarial networks and transfer
  learning for internet of vehicle,'' \emph{EURASIP Journal on Advances in
  Signal Processing}, vol. 2021, no.~1, pp. 1--19, 2021.

\bibitem{arjovsky2017wasserstein}
M.~Arjovsky, S.~Chintala, and L.~Bottou, ``Wasserstein generative adversarial
  networks,'' in \emph{International conference on machine learning}.\hskip 1em
  plus 0.5em minus 0.4em\relax PMLR, 2017, pp. 214--223.

\bibitem{johnson2016perceptual}
J.~Johnson, A.~Alahi, and L.~Fei-Fei, ``Perceptual losses for real-time style
  transfer and super-resolution,'' in \emph{European conference on computer
  vision}.\hskip 1em plus 0.5em minus 0.4em\relax Springer, 2016, pp. 694--711.

\bibitem{ledig2017photo}
C.~Ledig, L.~Theis, F.~Husz{\'a}r, J.~Caballero, A.~Cunningham, A.~Acosta,
  A.~Aitken, A.~Tejani, J.~Totz, Z.~Wang \emph{et~al.}, ``Photo-realistic
  single image super-resolution using a generative adversarial network,'' in
  \emph{Proceedings of the IEEE conference on computer vision and pattern
  recognition}, 2017, pp. 4681--4690.

\bibitem{kupyn2018deblurgan}
O.~Kupyn, V.~Budzan, M.~Mykhailych, D.~Mishkin, and J.~Matas, ``Deblurgan:
  Blind motion deblurring using conditional adversarial networks,'' in
  \emph{Proceedings of the IEEE conference on computer vision and pattern
  recognition}, 2018, pp. 8183--8192.

\bibitem{alam2021survey}
M.~S. Alam and J.~Oluoch, ``A survey of safe landing zone detection techniques
  for autonomous unmanned aerial vehicles (uavs),'' \emph{Expert Systems with
  Applications}, vol. 179, p. 115091, 2021.

\bibitem{haghighi2022performance}
H.~Haghighi, D.~Delahaye, and D.~Asadi, ``Performance-based emergency landing
  trajectory planning applying meta-heuristic and dubins paths,'' \emph{Applied
  Soft Computing}, vol. 117, p. 108453, 2022.

\bibitem{lin2020fast}
Z.~Lin, L.~Castano, E.~Mortimer, and H.~Xu, ``Fast 3d collision avoidance
  algorithm for fixed wing uas,'' \emph{Journal of Intelligent \& Robotic
  Systems}, vol.~97, no.~3, pp. 577--604, 2020.

\bibitem{goss2004aircraft}
J.~Goss, R.~Rajvanshi, and K.~Subbarao, ``Aircraft conflict detection and
  resolution using mixed geometric and collision cone approaches,'' in
  \emph{AIAA Guidance, Navigation, and Control Conference and Exhibit}, 2004,
  p. 4879.

\bibitem{daniels2012collision}
Z.~Daniels, L.~Wright, J.~Holt, and S.~Biaz, ``Collision avoidance of multiple
  uas using a collision cone-based cost function,'' \emph{Computer Science and
  Software Engineering Department, Auburn University, Tech. Rep. CSSE12-07},
  2012.

\bibitem{iiyama2021deep}
K.~Iiyama, K.~Tomita, B.~A. Jagatia, T.~Nakagawa, and K.~Ho, ``Deep
  reinforcement learning for safe landing site selection with concurrent
  consideration of divert maneuvers,'' \emph{arXiv preprint arXiv:2102.12432},
  2021.

\bibitem{theodore2006flight}
C.~Theodore, D.~Rowley, A.~Ansar, L.~Matthies, S.~Goldberg, D.~Hubbard, and
  M.~Whalley, ``Flight trials of a rotorcraft unmanned aerial vehicle landing
  autonomously at unprepared sites,'' in \emph{Annual forum
  proceedings-American helicopter society}, vol.~62, no.~2.\hskip 1em plus
  0.5em minus 0.4em\relax AMERICAN HELICOPTER SOCIETY, INC, 2006, p. 1250.

\bibitem{edwards2007vision}
B.~Edwards, J.~Archibald, W.~Fife, and D.-J. Lee, ``A vision system for
  precision mav targeted landing,'' in \emph{2007 International Symposium on
  Computational Intelligence in Robotics and Automation}.\hskip 1em plus 0.5em
  minus 0.4em\relax IEEE, 2007, pp. 125--130.

\bibitem{Yang2020}
X.~Yang and P.~Wei, ``{Scalable multi-agent computational guidance with
  separation assurance for autonomous urban air mobility},'' \emph{Journal of
  Guidance, Control, and Dynamics}, vol.~43, no.~8, pp. 1473--1486, 2020.

\bibitem{Siewert2019}
S.~Siewert, K.~Sampigethaya, J.~Buchholz, and S.~Rizor, ``{Fail-Safe,
  Fail-Secure Experiments for Small UAS and UAM Traffic in Urban Airspace},''
  \emph{AIAA/IEEE Digital Avionics Systems Conference - Proceedings}, vol.
  2019-Septe, 2019.

\bibitem{ellis2020time}
K.~Ellis, J.~Koelling, M.~Davies, and P.~Krois, ``In-time system-wide safety
  assurance (issa) concept of operations and design considerations for urban
  air mobility (uam),'' 2020.

\bibitem{Bosson2018}
C.~S. Bosson and T.~A. Lauderdale, ``{Simulation evaluations of an autonomous
  urban air mobility network management and separation service},'' \emph{2018
  Aviation Technology, Integration, and Operations Conference}, 2018.

\bibitem{mathur2019paths}
A.~Mathur, K.~Panesar, J.~Kim, E.~M. Atkins, and N.~Sarter, ``Paths to
  autonomous vehicle operations for urban air mobility,'' in \emph{AIAA
  Aviation 2019 Forum}, 2019, p. 3255.

\bibitem{degas2022survey}
A.~Degas, M.~R. Islam, C.~Hurter, S.~Barua, H.~Rahman, M.~Poudel, D.~Ruscio,
  M.~U. Ahmed, S.~Begum, M.~A. Rahman \emph{et~al.}, ``A survey on artificial
  intelligence (ai) and explainable ai in air traffic management: Current
  trends and development with future research trajectory,'' \emph{Applied
  Sciences}, vol.~12, no.~3, p. 1295, 2022.

\bibitem{chen2018workload}
M.~Chen and D.~Zhu, ``A workload balanced algorithm for task assignment and
  path planning of inhomogeneous autonomous underwater vehicle system,''
  \emph{IEEE transactions on cognitive and developmental systems}, vol.~11,
  no.~4, pp. 483--493, 2018.

\bibitem{boots2009spatial}
B.~Boots, K.~Sugihara, S.~N. Chiu, and A.~Okabe, ``Spatial tessellations:
  concepts and applications of voronoi diagrams,'' 2009.

\bibitem{wu2011distributed}
C.~Wu, G.~S. Tewolde, W.~Sheng, B.~Xu, and Y.~Wang, ``Distributed
  multi-actuator control for workload balancing in wireless sensor and actuator
  networks,'' \emph{IEEE Transactions on Automatic Control}, vol.~56, no.~10,
  pp. 2462--2467, 2011.

\bibitem{xue2009airspace}
M.~Xue, ``Airspace sector redesign based on voronoi diagrams,'' \emph{Journal
  of Aerospace Computing, Information, and Communication}, vol.~6, no.~12, pp.
  624--634, 2009.

\bibitem{damani2021primal}
M.~Damani, Z.~Luo, E.~Wenzel, and G.~Sartoretti, ``Primal $ \_2 $: Pathfinding
  via reinforcement and imitation multi-agent learning-lifelong,'' \emph{IEEE
  Robotics and Automation Letters}, vol.~6, no.~2, pp. 2666--2673, 2021.

\bibitem{sartoretti2019primal}
G.~Sartoretti, J.~Kerr, Y.~Shi, G.~Wagner, T.~S. Kumar, S.~Koenig, and
  H.~Choset, ``Primal: Pathfinding via reinforcement and imitation multi-agent
  learning,'' \emph{IEEE Robotics and Automation Letters}, vol.~4, no.~3, pp.
  2378--2385, 2019.

\end{thebibliography}

\begin{IEEEbiographynophoto}{}

\begin{IEEEbiography}
[{\includegraphics[width=1in,height=1.25in,clip,keepaspectratio]{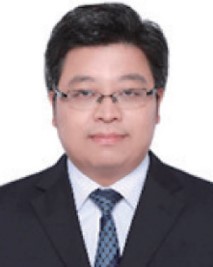}}]{Xinyu Zhang} was born in Huining, Gansu Province, and he received a B.E. degree from the School of Vehicle and Mobility at Tsinghua University, in 2001. He was a visiting scholar at the University of Cambridge.
\newline He is currently a researcher with the School of Vehicle and Mobility, and the head of the Mengshi Intelligent Vehicle Team at Tsinghua University. Dr. Zhang is the author of more than 30 SCI/EI articles. His research interests include intelligent driving and multimodal information fusion.
\end{IEEEbiography}

\begin{IEEEbiography}[{\includegraphics[width=1in,height=1.25in,clip,keepaspectratio]{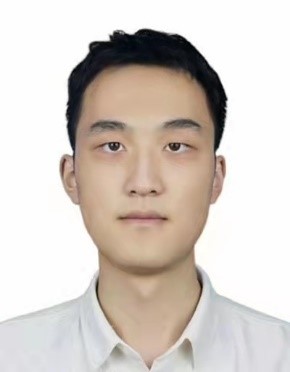}}]{Jiangeng Huang}
was born in Changchun, Jilin Province, and he received an M.Sc. degree from the School of Mechanism Engineering at the National University of Singapore, in 2021.
\newline He is currently a research assistant with the School of Vehicle and Mobility, Tsinghua University, and works in the Mengshi Intelligent Vehicle Team. His research interests include 3D object detection in computer vision and deep reinforcement learning for robotics.
\end{IEEEbiography}

\begin{IEEEbiography}[{\includegraphics[width=1in,height=1.25in,clip,keepaspectratio]{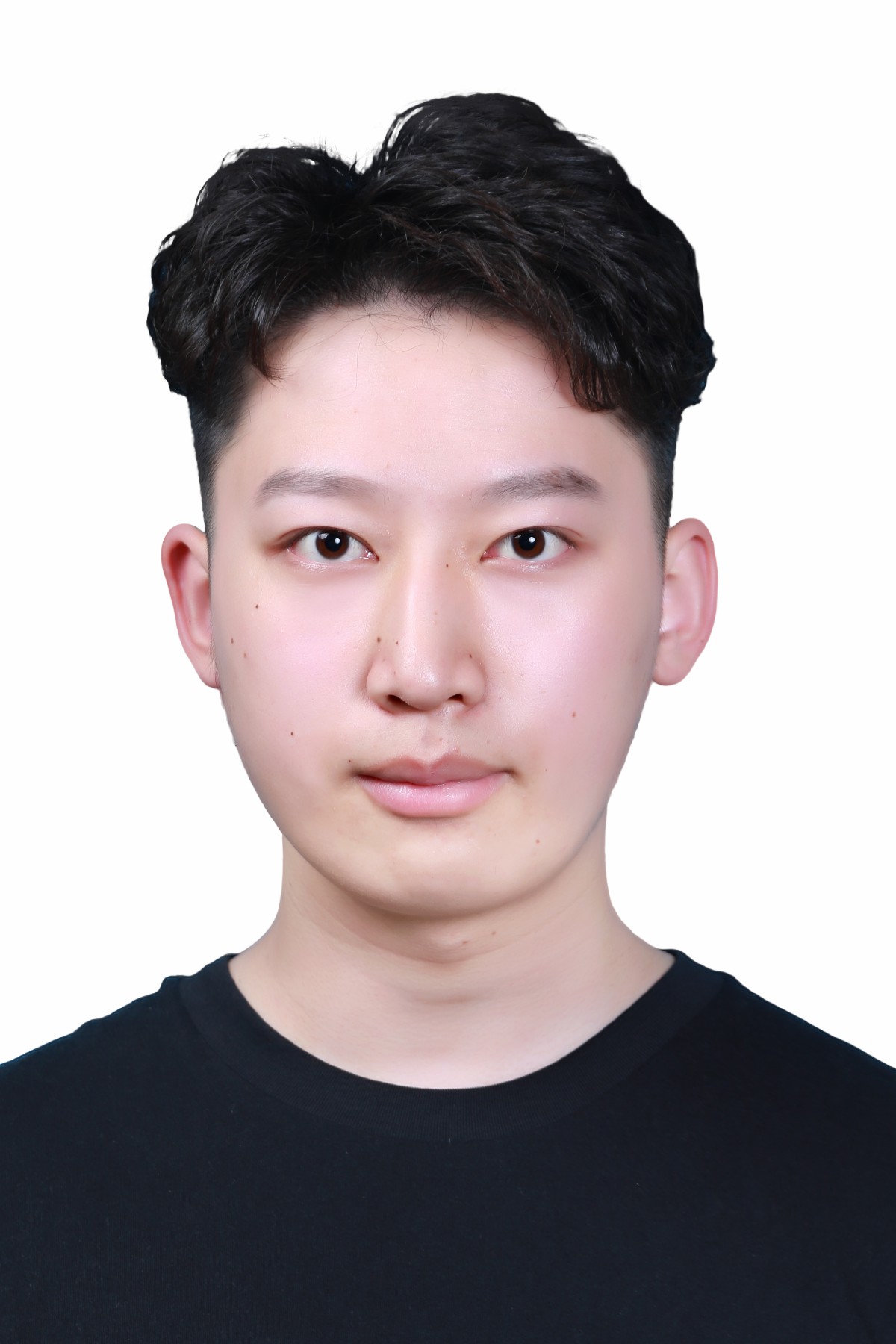}}]
{Yuanhao Huang} received the B.E. degree from the Institute of Disaster Prevention Department, Beijing, China, in 2020. Now he is pursuing a Master degree in engineering at Inner Mongolia University of Technology, Hohhot, China. 
He is currently working on a joint training program at New Technology Concept Automobile Research Institute, Tsinghua University. His research interests include robotics, automatic control and SLAM.
\end{IEEEbiography}

\begin{IEEEbiography}
[{\includegraphics[width=1in,height=1.25in,clip,keepaspectratio]{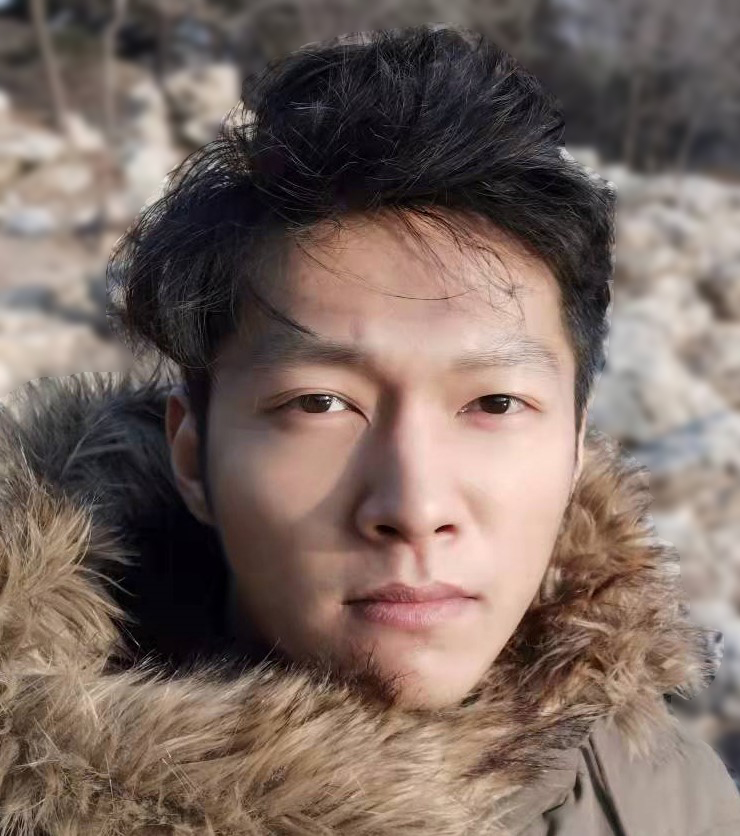}}]{Kangyao Huang} received the B.Eng. degree in Aerospace from Northwestern Polytechnical University, Xi’an, China, in 2016, and the M.Res. degree in Control \& Systems Engineering from the University of Sheffield, Sheffield, U.K., in 2020. Currently he is pursuing a Ph.D degree at the Department of Computer Science and Technology, Tsinghua University, Beijing, China, and working with the Mengshi Intelligent Vehicle Team.
\newline He has three years working experience in aerospace industry. He was the Founder, Technical Director and Chief Engineer with Bingo Intelligence Aviation Technology co., LTD, where he developed the general software architecture for integrated avionics system, and provided applied research in cooperation with partners in aerospace, defense technology, and manufacturing sectors. His research interests include swarm robotics, unmanned aerial vehicle, aerospace engineering, cooperative behaviour and evolutionary algorithm.
\end{IEEEbiography}

\begin{IEEEbiography}
[{\includegraphics[width=1in,height=1.25in,clip,keepaspectratio]{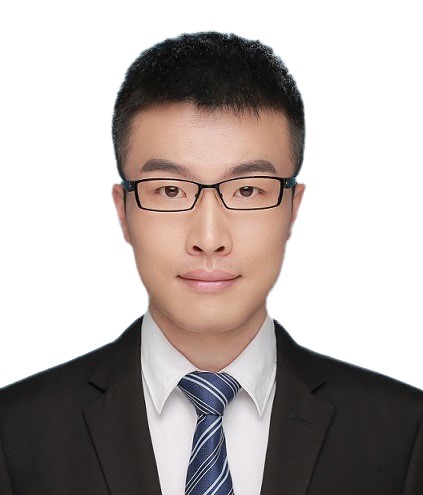}}]{Lei Yang} received the M.S. degree in mechanical engineering at robotics institute, Beihang University, in 2018. He is currently pursuing the Ph.D. degree in the State Key Laboratory of Automotive Safety and Energy, and the School of Vehicle and Mobility, Tsinghua University. His research interests include deep learning, vision-based 3D scene understanding and autonomous driving.
\end{IEEEbiography}

\begin{IEEEbiography}[{\includegraphics[width=1in,height=1.25in,clip,keepaspectratio]{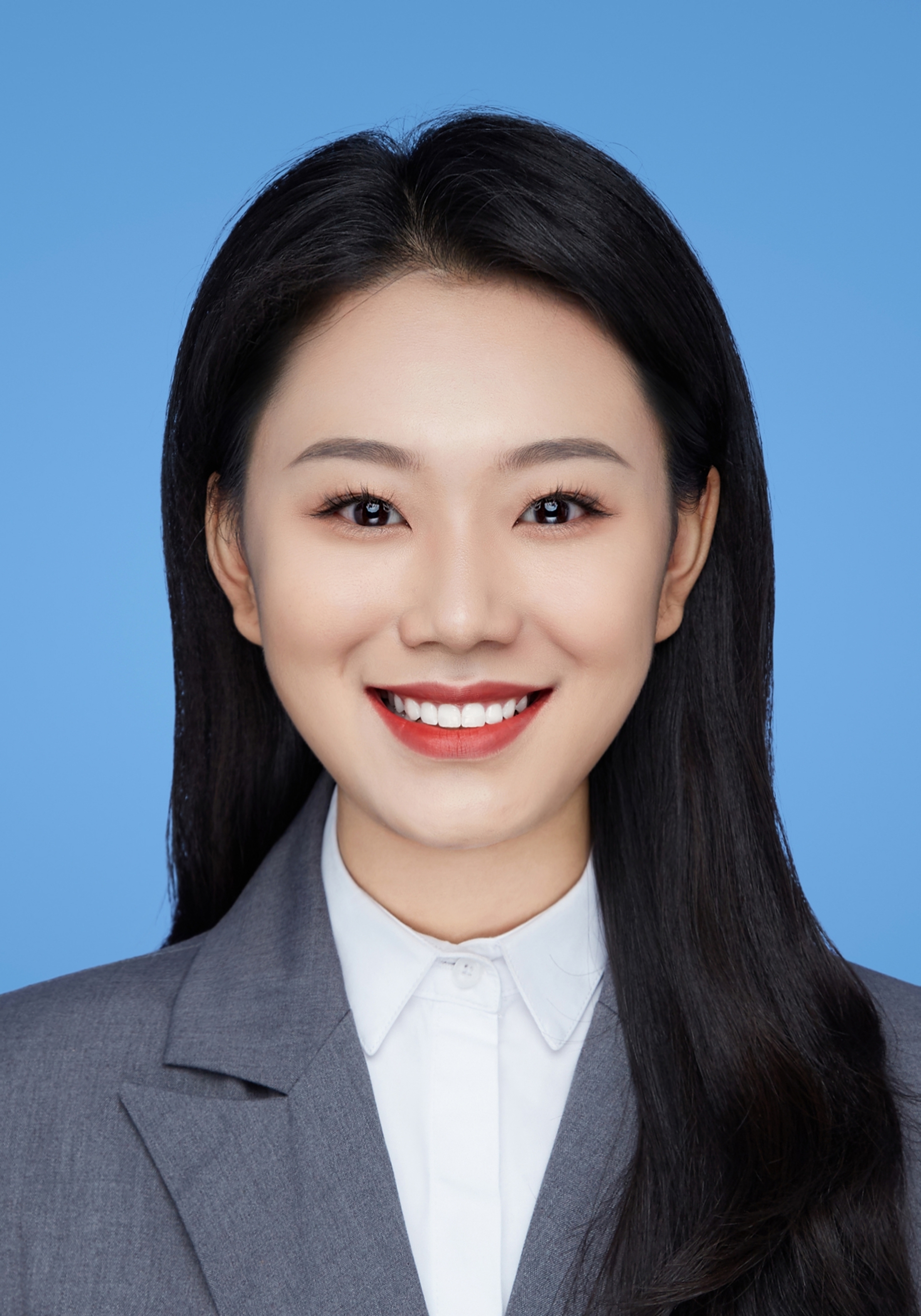}}]
{Han Yan} was born in Xing'an League, Inner Mongolia Autonomous Region. She received the B.S. degree in industrial design from North Minzu University, in 2019. She is currently pursuing the master’s degree at Beijing Forestry University.
\newline She is currently with the Mengshi Intelligent Vehicle Team, School of Vehicle and Mobility, Tsinghua University, Beijing, China. Her research interests include industrial design and service design.
\end{IEEEbiography}

\begin{IEEEbiography}
[{\includegraphics[width=1in,height=1.25in,clip,keepaspectratio]{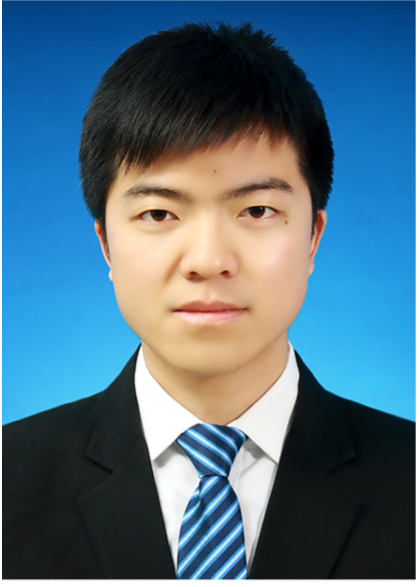}}]{Li Wang} was born in Shangqiu, Henan Province, China in 1990. He received his Ph.D. degree in mechatronic engineering at State Key Laboratory of Robotics and System, Harbin Institute of Technology, in 2020. He was a visiting scholar at Nanyang Technology of University for two years. Currently, he is a postdoctoral fellow in the State Key Laboratory of Automotive Safety and Energy, and the School of Vehicle and Mobility, Tsinghua University.
\newline Dr. Wang is the author of more than 15 SCI/EI articles. His research interests include autonomous driving perception, 3D robot vision and multi-modal fusion.
\end{IEEEbiography}

\begin{IEEEbiography}
[{\includegraphics[width=1in,height=1.25in,clip,keepaspectratio]{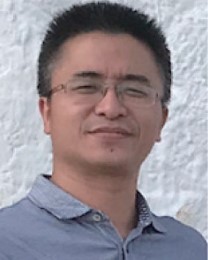}}]{Huaping Liu} (Senior Member, IEEE) is an Associate Professor with the Department of Computer Science and Technology, Tsinghua University, Beijing, China. His research interests include robot perception and learning. Dr. Liu has served as an Associate Editor of ICRA and IROS and in the Program Committees of IJCAI, RSS, and IJCNN. He is an Associate Editor of the IEEE ROBOTICS AND AUTOMATION LETTERS, Neurocomputing, and Cognitive Computation.
\end{IEEEbiography}

\begin{IEEEbiography}
[{\includegraphics[width=1in,height=1.25in,clip,keepaspectratio]{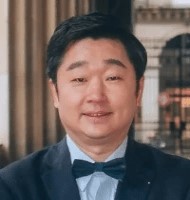}}]{Jianxi Luo}  holds a PhD in Engineering Systems from Massachusetts Institute of Technology, and M.S. and B.E. in Engineering from Tsinghua University. He is the Founder and Director of Data-Driven Innovation Lab at Singapore University of Technology and Design (SUTD). He teaches topics on engineering design, entrepreneurship, and innovation. His research interests include data-driven innovation and artificial intelligence for engineering design. 
\newline Prof. Luo is the Department Editor of the IEEE Transactions on Engineering Management, an Associate Editor of Design Science, Associate Editor of Artificial Intelligence for Engineering Design, Analysis \& Manufacturing, and Editorial Board Member of Research in Engineering Design, etc. He served as the Chair of INFORMS Technology Innovation Management \& Entrepreneurship Section.
\end{IEEEbiography}

\begin{IEEEbiography}
[{\includegraphics[width=1in,height=1.25in,clip,keepaspectratio]{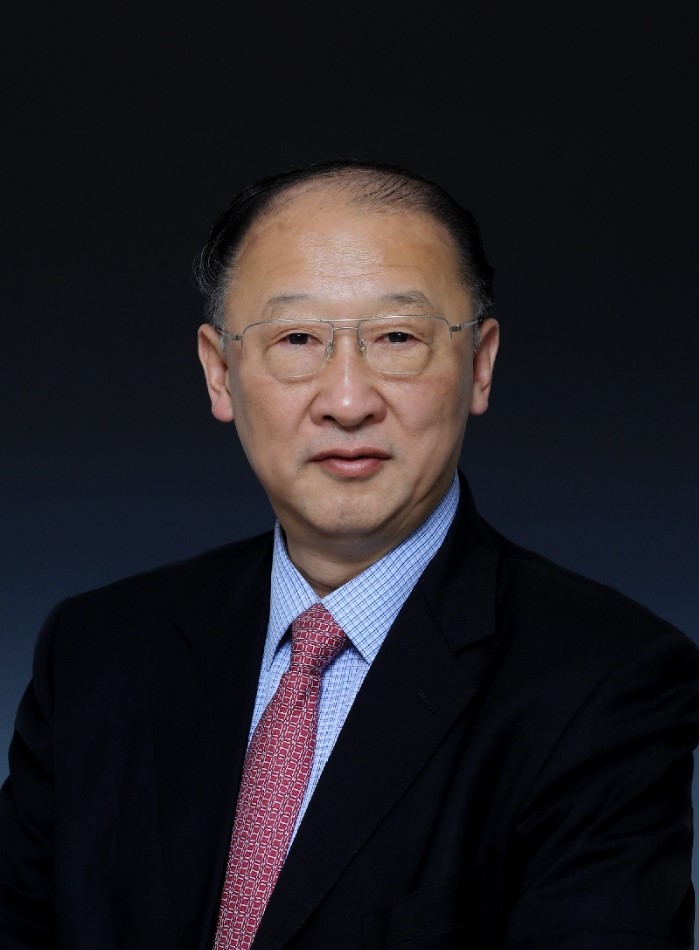}}]{Jun Li} received the Ph.D. degree in internal combustion engineering from Jilin Polytechnic University, Changchun, China, in 1989. He is a fellow of the Chinese Academy of Engineering, and the Vice-Chief Engineer and the Director of the Research and Development Center with China FAW Group Corporation, Changchun.
\end{IEEEbiography}

\end{IEEEbiographynophoto}

\end{document}